\RequirePackage{fix-cm}
\documentclass[smallextended]{svjour3}      
\smartqed  
\usepackage{graphicx}
\usepackage{mathptmx}      
\usepackage{color}
\usepackage{makecell}
\usepackage{lscape}
\usepackage{booktabs}
\usepackage{array}
\usepackage{multirow}

\begin{document}

\title{Optical Flow for Video Super-Resolution: A Survey}

\author{Zhigang Tu$^1$     \and
        Hongyan Li$^{2*}$     \and
        Wei Xie$^{3*}$        \and
        Yuanzhong Liu$^1$  \and
        Shifu Zhang$^4$    \and
        Baoxin Li$^5$      \and
        Junsong Yuan$^6$
}

\institute{Zhigang Tu and Yuanzhong Liu \at
              State Key Laboratory of Information Engineering in Surveying, Mapping and Remote Sensing, Wuhan University \\
              \email{tuzhigang@whu.edu.cn}           
           \and
           Corresponding author*: Hongyan Li \at
              School of Information Engineering, Hubei University of Economics, Wuhan \\
              \email{hongyanli2000@126.com.}
           \and
           Corresponding author*: Wei Xie \at
              School of Computer, Central China Normal University, Wuhan  \\
              \email{XW@mail.ccnu.edu.cn.}
           \and
           Baoxin Li \at
              School of Computing, Informatics, Decision System Engineering, Arizona State University  \\
              \email{baoxin.li@asu.edu}
           \and
           Junsong Yuan \at
              Computer Science and Engineering department, State University of New York at Buffalo  \\
              \email{jsyuan@buffalo.edu}
}

\date{Received: date / Accepted: date}

\maketitle

\begin{abstract}
Video super-resolution is currently one of the most active research topics in computer vision as it plays an important role in many visual applications. Generally, video super-resolution contains a significant component, i.e., motion compensation, which is used to estimate the displacement between successive video frames for temporal alignment. Optical flow, which can supply dense and sub-pixel motion between consecutive frames, is among the most common ways for this task. To obtain a good understanding of the effect that optical flow acts in video super-resolution, in this work, we conduct a comprehensive review on this subject for the first time. This investigation covers the following major topics: the function of super-resolution (i.e., why we require super-resolution); the concept of video super-resolution (i.e., what is video super-resolution); the description of evaluation metrics (i.e., how (video) super-resolution performs); the introduction of optical flow based video super-resolution; the investigation of using optical flow to capture temporal dependency for video super-resolution. Prominently, we give an in-depth study of the deep learning based video super-resolution method, where some representative algorithms are analyzed and compared. Additionally, we highlight some promising research directions and open issues that should be further addressed.

\keywords{Video super-resolution \and Optical flow \and Optical Flow-based video super-resolution \and Temporal dependency}
\end{abstract}

\section{Introduction}
\label{intro}
Image resolution reflects the visual details viewable in an image, thus typically, images with higher resolution capture more visual information than the low resolution ones for both human perception and machine interpretation (\textcolor{blue}{Park et al. 2003; Fookes et al. 2004; Nguyen et al. 2018}). That is to say, higher resolution images supply clearer and more discriminative pictorial information for human to perceive, and finer details for machine to interpret (\textcolor{blue}{Xiong et al. 2010; Singh and Singh 2020}). Consequently, capturing high resolution (HR) images is extremely important for both human and machines (\textcolor{blue}{Park et al. 2003; Ledig et al. 2017a}).

However, low resolution (LR) images are widespread in the real world due to two main reasons related to the image sensors:
\begin{itemize}
\item The cost of an imaging sensor and its spatial resolution is directly related. Generally, an HR sensor is more costly than its LR counterpart, leading to the wide deployment of LR sensors in cost-sensitive applications. 
\item The spatial resolution is constrained by image sensing technology. Increasing the resolution may typically cause low signal-to-noise ratio (SNR) as less light can fall upon each sensor cell.
\end{itemize}

Given that many images are acquired by LR cameras, it is imperative to find a way for reconstructing HR images from captured LR images. This process is referred to as ``super-resolution (SR) image reconstruction" (\textcolor{blue}{Farsiu et al. 2004; Shi et al. 2016}). It has been an important and challenging technique in the field of computer vision and image processing, and continues to attract attention from the research community (\textcolor{blue}{Nasrollahi and Moeslund 2014; Thapa et al. 2016; Wang et al. 2021a}).

\begin{figure*}
\begin{center} 
 \includegraphics[width=1.0\linewidth]{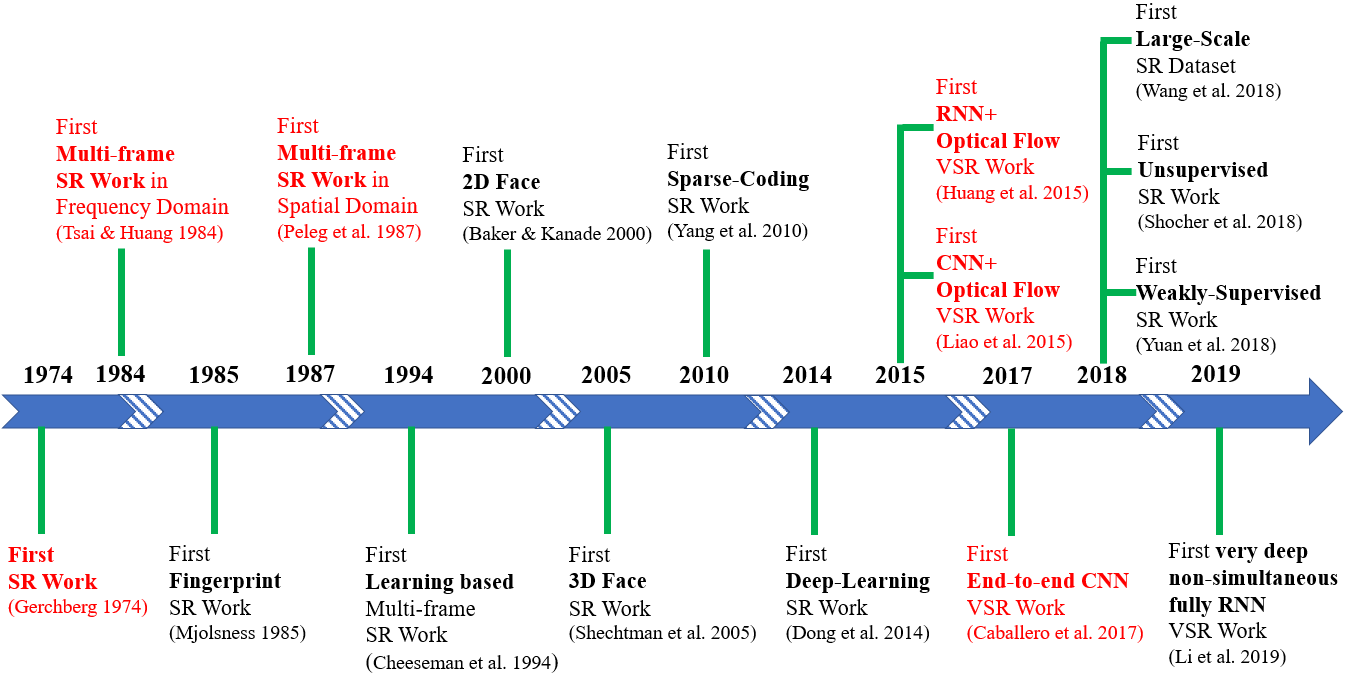}
 \caption{Timeline of the landmark work in the field of super-resolution.}
 \label{fig:Timeline}
\end{center}
\end{figure*}

Since the heuristic work of Tsai and Huang in 1984, SR has witnessed significant progresses (see Figure~\ref{fig:Timeline}) while being widely used in many domains (\textcolor{blue}{Yuan et al. 2010; Borsoi et al. 2019}):
\begin{itemize}
\item Visual entertainment: High-resolution images supply more comfortable visual experience, which is a long-term pursuit for human (\textcolor{blue}{Shen et al. 2015}). Recently, with the emergence of new display technology, the high-definition television (HDTV, 1920$\times$1080) and more advanced ultra high definition television UHDTV (3840$\times$2048 or 4K, 7680$\times$4320 or 8k), will dominate the consumer market. There is an increasing requirement for using SR method to transform LR videos into HR versions for being enjoyed on HR devices (\textcolor{blue}{Kappeler et al. 2016; Liu and Sun 2014; Liu et al. 2017}).

\item Video surveillance: Super-resolution is desirably required in video surveillance as it can supply more powerful and more distinguishable visual cues for many visual tasks, e.g., face recognition (\textcolor{blue}{Mudunuri and Biswas 2016; Chen et al. 2018; Ma et al. 2020}), action recognition (\textcolor{blue}{Zhang et al. 2019}), pose estimation (\textcolor{blue}{Hong et al. 2015; Sun et al. 2019}), human activity recognition (\textcolor{blue}{Ryoo et al. 2017}), person re-identification (\textcolor{blue}{Jing et al. 2017}).

\item Multimedia image processing: Super-resolution is effective for improving image and scene details, thus it is widely used in multimedia image processing, e.g., image denosing (\textcolor{blue}{Cruz et al. 2018; Huang et al. 2019}), image inpainting (\textcolor{blue}{Meur et al. 2013}), image retrieval (\textcolor{blue}{Tan et al. 2018}), image classification (\textcolor{blue}{Cai et al. 2019}), semantic segmentation (\textcolor{blue}{Zhao et al. 2018; Wang et al. 2020a}), object detection (\textcolor{blue}{Girshick et al. 2016; Na and Fox 2017}), object recognition (\textcolor{blue}{Eekeren et al. 2010; Sajjadi et al. 2017; Noor et al. 2019}).

\item Other applications: Super-resolution plays an important role in some other domains, e.g., medical imaging (\textcolor{blue}{Huang et al. 2017; You et al. 2020}), remote sensing image processing (\textcolor{blue}{Tatem et al. 2001; Lei et al. 2020}), infrared imaging (\textcolor{blue}{Choi et al. 2011; Han et al. 2018}), biometrics (\textcolor{blue}{Huang et al. 2003; Yuan et al. 2009; Bian et al. 2017}).
\end{itemize}

The main contributions of this work are four-fold:
\begin{enumerate} 
\item Although some review papers about single image super-resolution have been published, but in the past decades, few important survey work on VSR has come off the press. We overview the definition, the application, and the landmark work of VSR, particularly with an emphasis on the optical flow based VSR;
\item We discuss the role and performance of optical flow in VSR systematically and comprehensively for the first time, and explain its principle;
\item We classify the traditional and recent optical flow based VSR techniques into three categories, and investigate the current deep learning (+ optical flow) based VSR algorithms. The advantages and limitations of each technique are summarized;
\item We analyze the challenges and open issues for both the optical flow based VSR and VSR. The new trends and promising directions are stated to supply an insightful guidance for the community.
\end{enumerate} 

\begin{figure}
 \centering
 \includegraphics[width=1.0\linewidth]{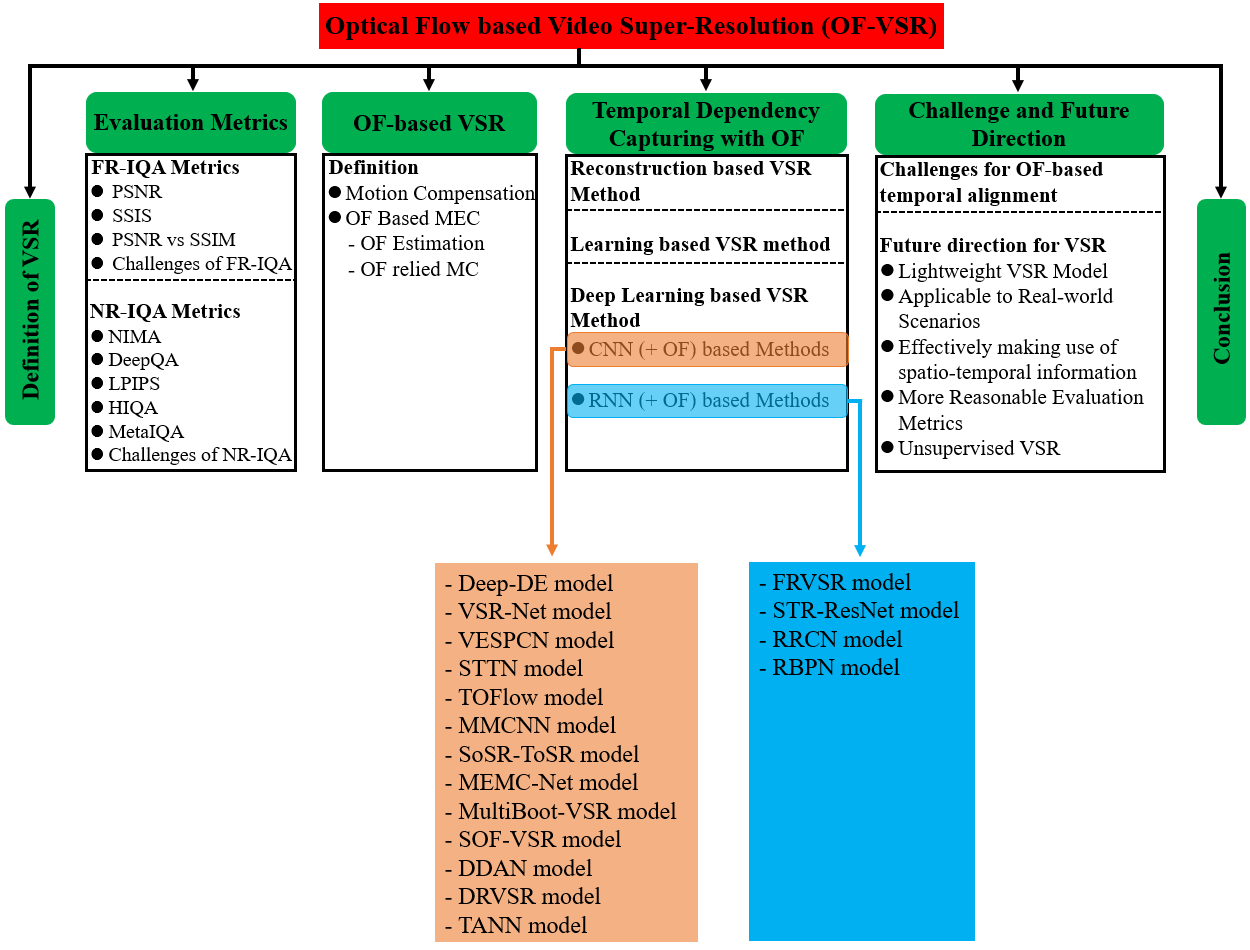}
 \caption{A hierarchical-structured taxonomy about OF-VSR of this survey.}
 \label{fig:Taxonomy}
\end{figure}

The rest of the paper is organized as follows. We introduce the concept of video super-resolution in Section 2. Section 3 describes the optical flow based video super-resolution technique. The methods to capture the temporal dependency via optical flow for VSR are discussed in Section 4, with more attention given to deep learning (+ optical flow) methods. We present remaining challenges and future directions in Section 5. Finally, the conclusion is given in Section 6. Figure~\ref{fig:Taxonomy} shows the overall taxonomy about optical flow based video super-resolution (OF-VSR) covered in this review paper in a hierarchical-structured framework.

\begin{figure*}
 \centering
 \includegraphics[width=1.0\linewidth]{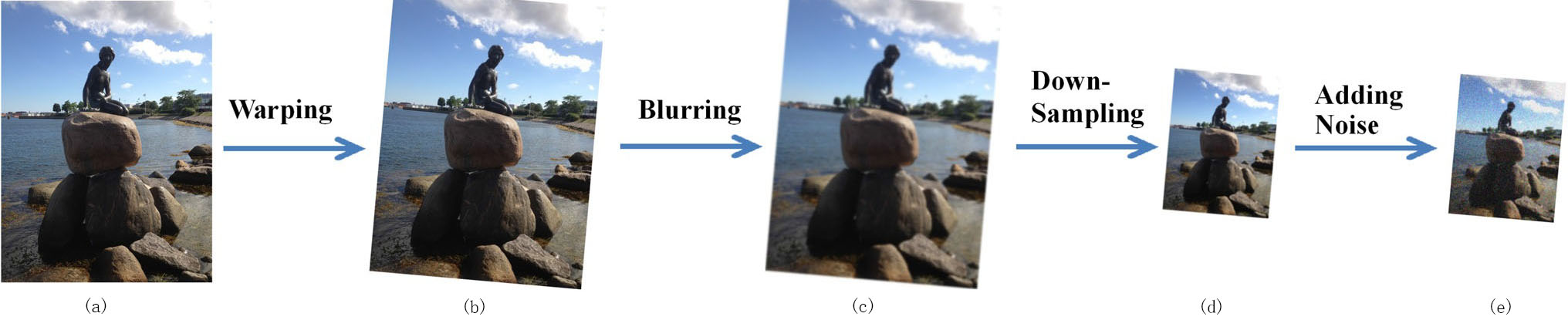}
 \caption{A typical sample of the modeling of the degradation process. From left to right, the degradation process contains warping, blurring, downsampling, and noise. (a) is a ground-truth HR video frame, and (e) is the corresponding degraded LR video frame of (a). The sample is taken from \textcolor{blue}{Nasrollahi and Moeslund 2014}.}
 \label{fig:Imaging-SR}
\end{figure*}

\section{What is Video Super-Resolution}
\label{sec:2}
Video (or multi-frame) super-resolution (VSR) is a technique that addresses the issue of how to reconstruct high resolution (HR) images with better visual quality and finer spectral details by combining complimentary information from multiple low-resolution (LR) counterparts (\textcolor{blue}{Peleg et al. 1987; Lin et al. 2005; Sajjadi et al. 2018; Daithankar and Ruikar 2020}). VSR derives from a natural phenomenon that LR images are subsampled and contain sub-pixel shifts, and thus the complementary information between LR images can be incorporated into a single image with higher resolution than the original observations (\textcolor{blue}{Picku 2007; Mudenagudi et al. 2011}). Enlarging the resolution can be treated as either improving the signal-to-noise ratio while preserving the size fixed, and/or depicting the image at a larger size with reasonable approximations for frequencies higher than those represented at the original size (\textcolor{blue}{Lin et al. 2005; Li et al. 2010}).

Mathematically, let $\textbf{I}$ represents a ground-truth HR video and $\textbf{I}_t$ represents a HR video frame at time $t$, $\textbf{F}_t$ denotes a degraded LR video frame. The degradation process of the HR video sequences can be expressed as follows:
\begin{equation}
  \textbf{F} = \textbf{D}(\textbf{I};\eta)
  \label{Def-1}
\end{equation}
where $\textbf{D}$ is the degradation function, $\eta$ represents the parameters of the degradation function which includes various degradation factors, e.g. noise, motion blur, scaling factor.

In practice, the degradation process (i.e. $\textbf{D}$ and $\eta$) is unknown, as only the LR video frames $\textbf{F}$ are given, but the degradation factors, which are quite complicated, are unknown. Accordingly, we need to restore a HR video approximation $\widehat{\textbf{I}}$ of the ground truth HR video $\textbf{I}$ from the LR video $\textbf{F}$, which can be formulated as:
\begin{equation}
  \widehat{\textbf{I}} = \textbf{S}(\textbf{F};\vartheta)
  \label{Def-2}
\end{equation}
where $\textbf{S}$ denotes the VSR model and $\vartheta$ represents the parameters of the model $\textbf{S}$.

The degradation process is usually affected by different factors (e.g., compression artifacts, anisotropic degradations, sensor noise, speckle noise) (\textcolor{blue}{Wang et al. 2021a}). Many attempts have been tried to simulate this process (\textcolor{blue}{Zhang et al. 2018a}), Figure~\ref{fig:Imaging-SR} shows a typical example of the degradation process that is assumed in most SR methods.
Currently, the widely adopted strategy is defined as:
\begin{equation}
  \textbf{D}(\textbf{I};\eta) = (\textbf{F} \otimes k)\downarrow_s + n_\zeta, \{s, k, \zeta\} \subset \eta
  \label{Def-DegPro}
\end{equation}
where $\downarrow_s$ represents a downsampling conduction which is performed with a scaling factor $s$, $\otimes$ is a convolution operation and $k$ is the blur kernel. $n_\zeta$ denotes some additive white Gaussian noise with a standard deviation $\zeta$.
(\textcolor{blue}{Wang et al. 2021a}) stated that this combinative degradation model Eq.(3) is closer to realistic conditions and brings more benefits to VSR.

Finally, the objective function of VSR can be formulated as following:
\begin{equation}
  \widehat{\vartheta} = \arg_{\vartheta} \min L(\textbf{I},\widehat{\textbf{I}}) + \lambda \Phi(\vartheta)
  \label{Def-VSR}
\end{equation}
where $L(\textbf{I},\widehat{\textbf{I}})$ denotes the loss function between the generated HR video $\widehat{\textbf{I}}$ and the ground truth video $\textbf{I}$. $\Phi(\vartheta)$ represents the regularization term and $\lambda$ is the tradeoff parameter.

From Eq.~(\ref{Def-VSR}), we can find that VSR is an inversion problem as the process relies on the determination of the HR video $\textbf{I}$ from multiple low resolution observations $\textbf{F}_t$.

\textit{\textbf{Summary}}. Both spatial information with a frame and temporal information across different frames are important and useful for VSR (\textcolor{blue}{Li et al. 2016; Isobe et al. 2020}).

\section{Evaluation Metrics}
Image quality assessment (IQA) refers to automatically evaluate the perceptual quality of a distorted image. IQA plays a crucial role in the low-level computer vision community due to it has a wide range of applications in image restoration, image retrieval, image quality monitoring systems, etc (\textcolor{blue}{Zhu et al. 2020; Zhai and Min 2020}).

Since IQA normally serves as a basis for video quality assessment (VQA) because of videos are sequences of images, IQA is always the primary research topic. For understanding more evaluation metrics about IQA and VQA, please refer to a most recently survey paper (\textcolor{blue}{Zhai and Min 2020}).

\subsection{Full-Reference IQA (FR-IQA) Metrics}
\subsubsection{Peak-Signal-to-Noise Ratio (PSNR)}
Peak-signal-to-noise ratio (PSNR), which is a crucial performance measure for lossy transformation, has long-term been widely used to evaluate the quality of both single image SR and VSR (\textcolor{blue}{Kim and Kwon 2010, Hore and Ziou 2010}). For SR, PSNR is calculated in terms of the maximum pixel value and the mean squared error (MSE) between the ground truth image $\textbf{I}_{Gt}$ and the restored image $\textbf{I}_t$:
\begin{equation}
  PSNR=10\cdot\lg(\frac{M^2}{\frac{1}{N}\sum^N_{i=1}(\textbf{I}_{Gt}(i)-\textbf{I}_t(i))^2})
  \label{PSNR}
\end{equation}
where $N$ represents the total number of pixels of the image \textit{e.g.} $\textbf{I}_{Gt}$, and $M$ denotes the maximum pixel value (normally $M = 255$), $i$ is a pixel position of the image. The $MSE$ is computed as:
\begin{equation}
  MSE=\frac{1}{N}\sum^N_{i=1}(\textbf{I}_{Gt}(i)-\textbf{I}_t(i))^2
  \label{MSE}
\end{equation}
The PSNR value approaches infinity if $\textbf{I}_{Gt}-\textbf{I}_t = 0$, \textit{i.e.} the super-resolved image $\textbf{I}_t$ is similar to the ground truth image $\textbf{I}_{Gt}$. Which means that a higher PSNR value accompany with a higher reconstructed image quality.

\vspace{0.3cm}
\textit{\textbf{Summary}}. The characteristics of PSNR can be summarized as following:
\begin{enumerate}
  \item Depending on the pixel-level MSE and focuses on the corresponding pixels' difference;
  \item Ignoring the human visual perception;
  \item Insensitively to distinguish the structural content of image since different types of degradations can generate the same value of MSE;
  \item Performing poor in complex real-world cases.
\end{enumerate}

\subsubsection{Structural Similarity Index Measure (SSIM)}
To capture high quality perception, (\textcolor{blue}{Wang et al. 2004}) proposed a structural similarity index measure (SSIM), which takes into account contrast, luminance distortion, and structure change between the two images. Since SSIM replaces the traditional error summation approaches (i.e. MSE) as PSNR with the structural similarity measuring, it can better simulate the human visual system (HVS) than PSNR. Specifically, SSIM is designed by integrating three factors :
\begin{equation}
  SSIM = [L(\textbf{I}_t, \textbf{I}_{Gt})]^\alpha[C(\textbf{I}_t, \textbf{I}_{Gt})]^\beta[S(\textbf{I}_t, \textbf{I}_{Gt})]^\gamma
  \label{SSIM}
\end{equation}
where the first factor $L(\textbf{I}_t, \textbf{I}_{Gt})$ is the luminance comparison function, the second factor $C(\textbf{I}_t, \textbf{I}_{Gt})$ is the contrast comparison function, and the third factor $S(\textbf{I}_t, \textbf{I}_{Gt})$ is the structure comparison function. $\alpha, \beta, \gamma$ are the weight parameters that used to adjust the relative importance of these three factors.

\noindent 1) $L(\textbf{I}_t, \textbf{I}_{Gt})$ evaluates the closeness of the two images’ mean luminance:
\begin{equation}
  L(\textbf{I}_t, \textbf{I}_{Gt}) = \frac{2\mu_{\textbf{I}_t}\mu_{\textbf{I}_{Gt}}+C_1}{\mu^2_{\textbf{I}_t}+\mu^2_{\textbf{I}_{Gt}}+C_1}
  \label{Luminance}
\end{equation}
where $\mu_{\textbf{I}_t}$ and $\mu_{\textbf{I}_{Gt}}$ respectively represents the mean luminance of $\textbf{I}_t$ and $\textbf{I}_{Gt}$. For instance, $\mu_{\textbf{I}_t}$ is computed as:
\begin{equation}
  \mu_{\textbf{I}_t} = \frac{1}{N}\sum^N_{i=1}\textbf{I}_t(i)
  \label{MLuminance}
\end{equation}
$L(\textbf{I}_t, \textbf{I}_{Gt})$ is maximal and equal to 1 if $\mu_{\textbf{I}_t} = \mu_{\textbf{I}_{Gt}}$.

\noindent 2) $C(\textbf{I}_t, \textbf{I}_{Gt})$ assesses the similarity of the two images’ contrast.
\begin{equation}
  C(\textbf{I}_t, \textbf{I}_{Gt}) = \frac{2\sigma_{\textbf{I}_t}\sigma_{\textbf{I}_{Gt}}+C_2}{\sigma^2_{\textbf{I}_t}+\sigma^2_{\textbf{I}_{Gt}}+C_2}
  \label{Contrast}
\end{equation}
where $\sigma_{\textbf{I}_t}$ and $\sigma_{\textbf{I}_{Gt}}$ separately denotes the standard deviation of $\textbf{I}_t$ and $\textbf{I}_{Gt}$. For example, $\sigma_{\textbf{I}_t}$ is evaluated as:
\begin{equation}
  \sigma_{\textbf{I}_t} = [\frac{1}{N-1}\sum^N_{i=1}(\textbf{I}_t(i)-\mu_{\textbf{I}_t})^2]^{\frac{1}{2}}
  \label{SContrast}
\end{equation}
$C(\textbf{I}_t, \textbf{I}_{Gt})$ is maximal and equal to 1 when $\sigma_{\textbf{I}_t}=\sigma_{\textbf{I}_{Gt}}$.

\noindent 3) $S(\textbf{I}_t, \textbf{I}_{Gt})$ is exploited to measure the correlation coefficient of the two images.

This is because the image structure can be expressed by the normalized pixel values (e.g. $\textbf{I}_t-\mu_{\textbf{I}_t}$), leading to their correlations are useful for reflecting the structural similarity. $S(\textbf{I}_t, \textbf{I}_{Gt})$ is calculated as:
\begin{equation}
  S(\textbf{I}_t, \textbf{I}_{Gt}) = \frac{\sigma_{\textbf{I}_t\textbf{I}_{Gt}}+C_3}{\sigma_{\textbf{I}_t}\sigma_{\textbf{I}_{Gt}}+C_3}
  \label{Structure}
\end{equation}
where $\sigma_{\textbf{I}_t\textbf{I}_{Gt}}$ is expressed as:
\begin{equation}
  \sigma_{\textbf{I}_t\textbf{I}_{Gt}} = \frac{1}{N-1}\sum^N_{i=1}(\textbf{I}_t(i)-\mu_{\textbf{I}_t})(\textbf{I}_{Gt}(i)-\mu_{\textbf{I}_{Gt}})
  \label{StructureSgm}
\end{equation}
$C_1, C_2, C_3$ are positive constants used for stability. $SSIM \subset [0, 1]$, $SSIM = 0$ means there is no correlation between the two images, while $SSIM = 1$ means the super-resolved image $\textbf{I}_t$ equals to the ground-truth image $\textbf{I}_{Gt}$.

\vspace{0.3cm}
\textit{\textbf{Summary}}. The characteristics of SSIM can be summarized as following:
\begin{enumerate}
  \item Suitable for evaluating the methods that without supplying sufficient texture details;
  \item Preferring blur over texture mismatching.
\end{enumerate}

\subsubsection{PSNR vs SSIM}
\begin{enumerate}
  \item SSIM and PSNR are more sensitive to noise degradation than the other degradations;
  \item PSNR is more sensitive to additive Gaussian noise, while SSIM is more sensitive to image compression.
\end{enumerate}

\subsubsection{Challenges of FR-IQA}
\begin{enumerate}
  \item FR-IQA measures unable to evaluate the perceptual quality precisely as they rely on the pixel-level error measures (e.g. L1 and L2 distances or their combination), leading to them focusing on pixel-level information locally;
  \item FR-IQA measures lack of generalization as they are formulated with limited and refined constraints, and required artificial intervention, causing them usually fail to model unknown distortions;
  \item FR-IQA measures are nearly unavailable in practice as they need non-distorted ground-truth reference images, but it is hard or impossible to obtain desired reference images in most cases.
\end{enumerate}

\subsection{No-Reference IQA (NR-IQA) Metrics: Deep Learning-Based Methods}
\subsubsection{NIMA}
(\textcolor{blue}{Talebi and Milanfar 2018}) explored a no-reference method, which called NIMA. It applies the CNN to predict the distribution of human opinion scores, leading to it can be trained on both the aesthetic and pixel-level quality datasets. Importantly, the NIMA method predicts the distribution of ratings to a histogram, to replace the conventional approaches that classify images to low/high score or regress to the mean score. As a result, a high correlation to human perception is gained. The squared earth mover’s distance (EMD) loss of (\textcolor{blue}{Hou et al. 2016}) is selected, as it improves the classification performance with ordered classes.

The normalized EMD based foss function is expressed as:
\begin{equation}
  EMD(p,\widehat{p}) = (\frac{1}{N}\sum_{k=1}^{N}|\textrm{CDF}_p(k)-\textrm{CDF}_{\widehat{p}}(k)|^r)^{1/r}
  \label{EMD}
\end{equation}
where $\textrm{CDF}_p(k)$ is a cumulative distribution function, which is computed as $\sum_{i=1}^{k}p_{_{Si}}$.
Moreover, the predicted quality probabilities are input to a soft-max function to ensure $\sum_{i=1}^{N}\widehat{p}_{_{Si}}=1$. Same as (\textcolor{blue}{Hou et al. 2016}), $r$ is set to 2 for optimization with gradient descent more convenient.

\subsubsection{DeepQA}
Inspired by the prior CNN-based visual-task approaches to learn the deep feature map, (\textcolor{blue}{Kim and Lee 2017}) proposed to use CNN to capture a visual sensitivity map, \textit{i.e.} a weighting map of reflecting the visual importance of each pixel to HVS. Accordingly a DeepQA model is formed, which in fact a CNN based full-reference image quality assessment (FR-IQA) model. In contrast to the traditional IQA measures, DeepQA aims to learning the optimal visual weight from the IQA dataset itself without requiring any prior knowledge of the HVS, where the training process requires some information of the dataset: a triplet of distorted images, objective error maps, and the subjective scores. Specifically:

1) The objective error map is defined as:
\begin{equation}
  e = \frac{\log(1/((\widehat{\textbf{I}}_R-\widehat{\textbf{I}}_D)^2+(\varepsilon/255^2)))}{\log(255^2/\varepsilon)}
  \label{OEM}
\end{equation}
where $\textbf{I}_R$ and $\textbf{I}_D$ respectively denotes the reference image and the distorted image, $\widehat{\textbf{I}}_R$ and $\widehat{\textbf{I}}_D$ are their normalized version.

2) The visual sensitivity map learned from CNN is defined as:
\begin{equation}
  s_1=CCN_1(\widehat{\textbf{I}}_D;\theta_1)
  \label{CNN-1}
\end{equation}
\begin{equation}
  s_2=CCN_2(\widehat{\textbf{I}}_D,e;\theta_2)
  \label{CNN-2}
\end{equation}
where $\theta_1$ and $\theta_2$ are the parameters of DeepQA. The perceptual error map is estimated by:
\begin{equation}
  p = s\odot e
  \label{PEM}
\end{equation}
where $\odot$ is the Hadamard product, $s$ is $s_1$ or $s_2$.

Accordingly, the pooled score is computed by averaging the cropped perceptual error map:
\begin{equation}
  \mu_p = \frac{1}{(H-8)(W-8)}\sum_{(i,j)\in \omega}p
  \label{PEM}
\end{equation}
where $H$ and $W$ are the height and width of $p$, $(i,j)$ is pixel index, and $\omega$ denotes the cropped region.

3) The final objective function of the DeepQA model is expressed as:
\begin{equation}
  {\L}_s(\widehat{\textbf{I}}_D;\theta) = \|(f(\mu_p)-S)\|^2_F
  \label{PEM}
\end{equation}
where $f(\cdot)$ denotes a nonlinear regression function, and $S$ represents the subjective score of the input distorted image.

\begin{figure*}[!htp]
 \centering
 \includegraphics[width=1.0\linewidth]{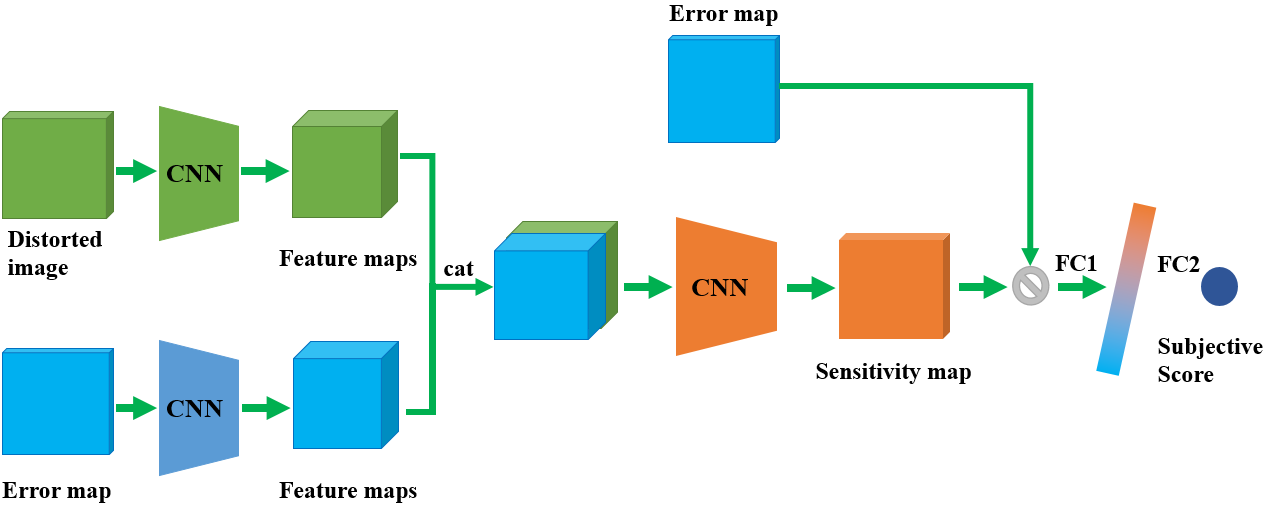}
 \caption{The architecture of the DeepQA model. Input: A distorted image and An error map, then producing a sensitivity map. Output: a subjective score, which is regressed by multiplying with the error map.}
 \label{fig:DeepQA}
\end{figure*}

The structure of DeepQA is described in Figure~\ref{fig:DeepQA}.

\vspace{0.3cm}
\textit{\textbf{Summary}}. The characteristics of DeepQA can be summarized as following:
\begin{itemize}
\item The DeepQA model, which with the usage of a triplet of a distorted image, its objective error map, and the subjective score, can capture the human visual sensitivity without any prior knowledge.
\item A total variation regularization, which penalizes the high frequency components of the estimated sensitivity map, enables the sensitivity map to more visually plausible without reducing the performance.
\item The model, which is optimized in an end-to-end manner, obtains the state-of-the-art correlation with human subjective scores.
\end{itemize}

\subsubsection{LPIPS}
(\textcolor{blue}{Zhang et al. 2018b}) construct a large-scale dataset for human perceptual similarity evaluation. Importantly, they assess the perceptual image patch similarity (LPIPS) by comparing the deep features with the classical metrics, where the deep features are learnt by CNNs with different architectures and visual tasks. The experimental results reveal that the deep features can model the perceptual similarity much better than the prior metrics that without CNNs.

The key component of LPIPS is the \textbf{\textit{network activations to distance}}, which can be expressed as:
\begin{equation}
  D(p, p_0) = \sum_l\frac{1}{H_lW_l}\sum_{h,w}\|w_l\odot(\widehat{F}^l_{hw}-\widehat{F}^l_{0hw})\|^2_2
  \label{EMD}
\end{equation}
where $D(p, p_0)$ represents the cosine distance between the reference and the distorted patches $p$ and $p_0$. $l$ denotes a convolution layer. $w$ is a vector used to scale a channel. $h$ represents the predict perceptual judgment. $\widehat{F}$ denotes the extracted deep feature, and $\widehat{F}^l, \widehat{F}^l_0\in \Re^{H_l\times W_l \times C_l}$ for layer $l$. (\textcolor{blue}{Zhang et al. 2018b}) scale the activations channel-wise via vector $w^l \in \Re^{C_l}$ and calculate the $l_2$ distance.

\vspace{0.3cm}
\textit{\textbf{Summary}}. The characteristics of LPIPS can be summarized as following:
\begin{itemize}
\item The stronger a feature set is at classification and detection, the stronger it is as a model of perceptual similarity judgments.
\item A good feature is a good feature. Features that are good at semantic tasks also provide good models of both human perceptual behavior
and macaque neural activity.
\end{itemize}

\subsubsection{Hallucinated-IQA (HIQA)}
(\textcolor{blue}{Lin and Wang 2018}) presented a hallucination-guided quality regression network (named HIQA), which takes the perceptual discrepancy into a deep neural network for learning, to imitate HVS and defeat the ill-posed nature of NR-IQA. Specifically, by making use of the perceptual discrepancy between the distorted image and the hallucinated reference, HIQA achieves an accurate and robust perceptual prediction.

\begin{figure*}[!htp]
 \centering
 \includegraphics[width=1.0\linewidth]{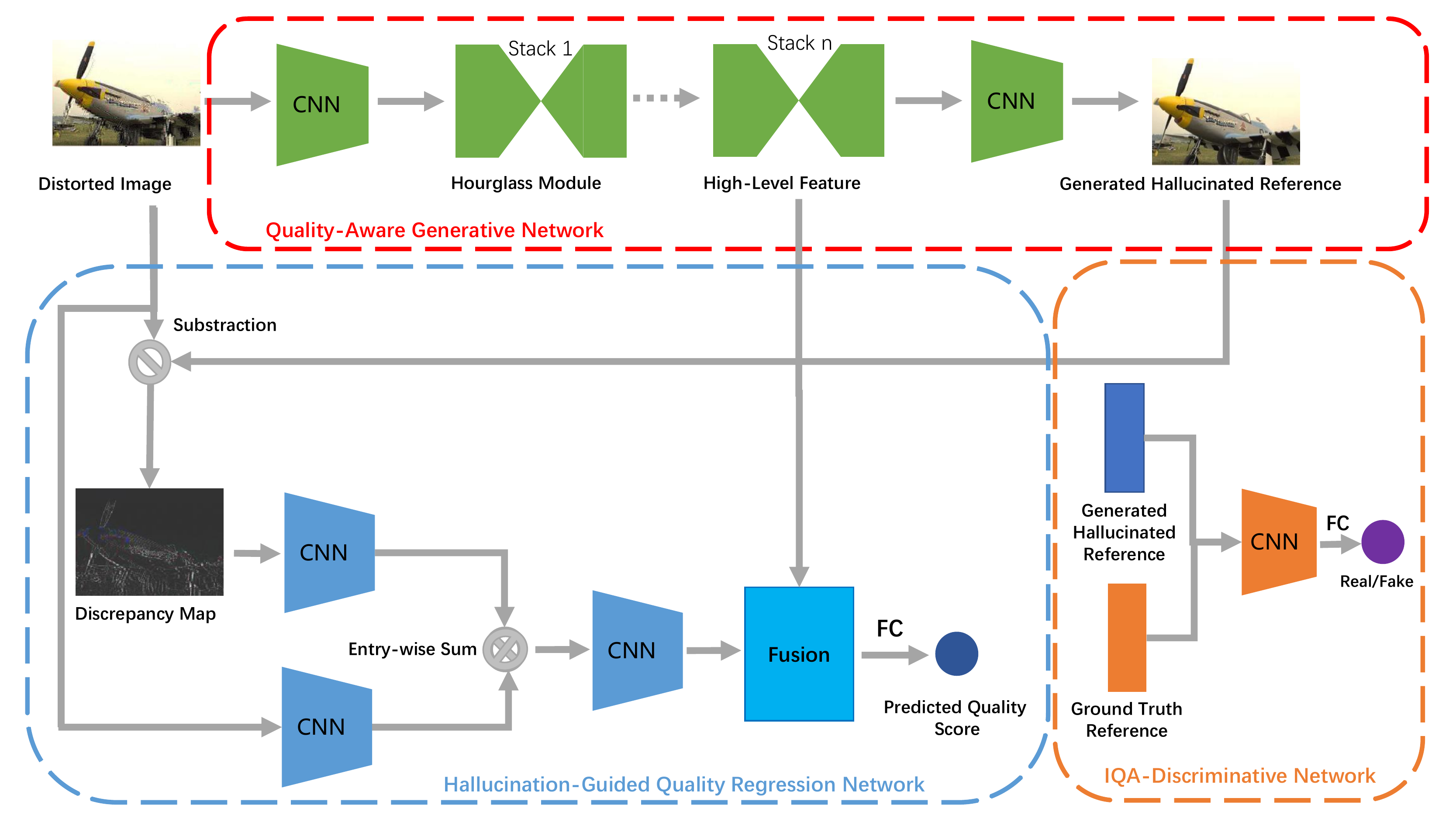}
 \caption{The architecture of the Hallucinated-IQA (HIQA) framework. There are three main components: (a) A Quality-Aware
Generative Network $G$, which is utilized to produce hallucinated reference images; (b) A Hallucination-Guided Quality Regression Network $R$, which is used to incorporate the discrepancy information between the hallucinated image and the distorted image encoded in the discrepancy map; (c) A IQA-Discriminator $D$, which is applied to refine the hallucinated image further.}
 \label{fig:HIQA}
\end{figure*}

As shown in Figure~\ref{fig:HIQA}, HIQA composes of three heavily related subnets.

\vspace{0.3cm}
\noindent \textbf{A. Quality Aware Generative Network}

The exploited Quality-Aware Generative Network aims to produce hallucinated reference images, where the hallucinated reference image is used to compensate the absence of true reference image. The gap between the hallucinated reference and the true reference is less, the performance of the quality regression network is better.

Importantly, to obtain high quality hallucinated reference images, HIQA exploits a quality-aware perceptual loss, which is able to incorporate the deep features of the regression network dynamically. The loss function is defined as:
\begin{equation}
  L_s(G_\theta(I^i_d),I^i_r)=\lambda_1L_v(G_\theta(I^i_d),I^i_r)+\lambda_2L_q(G_\theta(I^i_d),I^i_r)
  \label{QAPL}
\end{equation}
where
\begin{equation}
  L_v=\sum^{C_v}_{C_v=1}\frac{1}{W_jH_j}\sum^{W_j}_{x=1}\sum^{H_j}_{y=1}\|\phi_j(G_\theta(I^i_d))_{x,y}-\phi_j(I^i_r)_{x,y}\|^2
  \label{QAPL-Lv}
\end{equation}
and
\begin{equation}
  L_q=\sum^{C_q}_{C_q=1}\frac{1}{W_kH_k}\sum^{W_k}_{x=1}\sum^{H_k}_{y=1}\|\pi_k(G_\theta(I^i_d))_{x,y}-\pi_k(I^i_r)_{x,y}\|^2
  \label{QAPL-Lq}
\end{equation}
Particularly, $I_d$ is the distorted image, and $G(I_d)$ is the generating function; $I^i_d, \{i = 1, 2, ..., N\}$ denotes a set of distorted images, and $I^i_r, \{i = 1, 2, ..., N\}$ denotes the corresponding true reference images; $\phi_j(\cdot)$ represents the feature map at the $j-th$ layer of VGG-19, and $\pi_k(\cdot)$ represents the feature map at the $k-th$ layer of the hallucination-guided quality regression network $R$; $W$ and $H$ denote the dimensions of the feature map, $C$ is the number of feature maps at a particular layer.

\vspace{0.3cm}
\noindent \textbf{B. IQA-Discriminative Network}

IQA-Discriminator $D$ aims to reduce the affect of bad hallucination images to the deep regression network $R$, which is realized by distinguishing the fake samples from the real samples based on their positive or negative impact to $R$.

$G$ is optimized to fool the IQA-discriminator $D$ by producing the qualified hallucination scene to help improve $R$, and the adversarial loss of $G$ can be expressed as:
\begin{equation}
  L_{adv}=\textbf{E}[\log(1-D_\omega(G_\theta(I_d)))]
  \label{IQA-D}
\end{equation}
The overall loss function of $G$ for all training samples is defined by
\begin{equation}
  L_G=\mu_1L_p+\mu_2L_s+\mu_3L_{adv}
  \label{IQA-D-AL}
\end{equation}
where $\mu_1, \mu_2, \mu_3$ are the parameters which are aiming at keeping the trade off between the three loss components.

\vspace{0.3cm}
\noindent \textbf{C. Hallucination Guided Quality Regression Network}

The Hallucination-Guided Quality Regression Network $R$, which incorporates the discrepancy information with the high-level semantic fusion from the generative network $G$, to provide itself more plentiful and valid information. This is very useful to guide the network training.

\textbf{\textit{Discrepancy Map}}. (\textcolor{blue}{Lin and Wang 2018}) treat the distorted images and their discrepancy
maps as pairs $\{I^i_d, I^i_{map}\}^N_{i=1}$, a deep regression network can be trained according to:
\begin{equation}
  \widehat{\gamma}=\arg \min_\gamma\frac{1}{N}\sum^N_{i=1}L_r(R(I^i_d,I^i_{map}),s^i)
  \label{DMap}
\end{equation}
where the discrepancy map is defined as:
\begin{equation}
  I_{map} = |I_d - G_{\widehat{\theta}}(I_d)|
  \label{IMap}
\end{equation}

\textbf{\textit{High-level Semantic Fusion}}. The fusion term is defined as:
\begin{equation}
  F=f(H_{5,2}(I_d))\otimes(R_1(I_d,I_{map}))
  \label{HLSF}
\end{equation}
where $f$ is a linear projection to make the dimensions of $H$ and $R_1$ equally, $R_1$ represents the feature extraction before the fully connected layers ($R_2$) of $R$, and $\otimes$ represents concatenation.

\textbf{\textit{The Overall Loss Function}}. The final loss of $R$ is expressed as:
\begin{equation}
  L_R= \frac{1}{T}\sum^T_{t=1}\|R_2(f(H_{5,2(I_d)}))\otimes R_1(I_d,I_{map}) - s^t\|_{\ell_1}
  \label{DMap}
\end{equation}

\subsubsection{MetaIQA}
\begin{figure*}[!htp]
 \centering
 \includegraphics[width=0.6\linewidth]{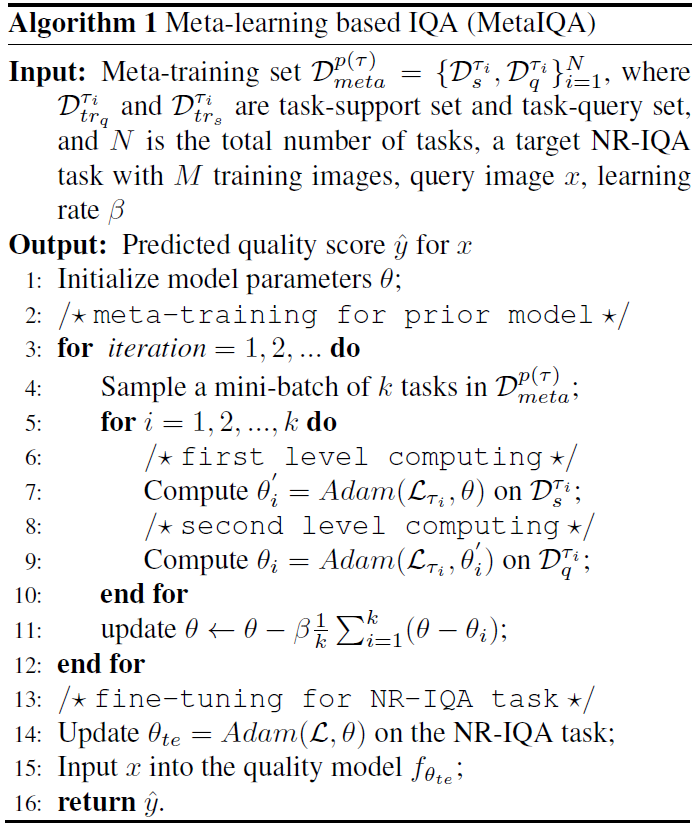}
 \caption{The entire procedure of Meta-learning based IQA (MetaIQA). The procedure of Algorithm 1 is copied from (\textcolor{blue}{Zhu et al. 2020}).}
 \label{fig:MetaIQA}
\end{figure*}

To model the human beings’ ability that ``getting the quality prior knowledge from images with various distortions easily and adapting to evaluate unknown distorted images quickly", (\textcolor{blue}{Zhu et al. 2020}) proposed a NR image quality metric MetaIQA based on the deep meta-learning, in which MetaIQA can capture the meta-knowledge shared by human when to assess the quality of images with multifarious distortions. In brief, the exploited MetaIQA enables the machines learn to learn, that is, to obtain the capacity of learning quickly from a relatively small amount of training samples for a related new task. Figure~\ref{fig:MetaIQA} shows the entire procedure of MetaIQA which is summarized in Algorithm 1.

\subsubsection{Challenges of NR-IQA}
\begin{enumerate}
  \item \textbf{The definition is ill-posed}: NR-IQA generally takes the distorted image as input for evaluation without any additional data. However, this scheme is counter-intuitive, as HVS requires a reference to measure the perceptual discrepancy by comparing the distorted image either directly with the original undistorted image or implicitly with a hallucinated scene in mind (\textcolor{blue}{Lin and Wang 2018});
  \item \textbf{The generalization ability is limited}: The deep-learning based NR-IQA metrics usually depend on pre-trained networks, however, these pre-trained networks are not designed for IQA, leading to their generalization performance is unsatisfactory when assessing unknown types of distortions.
\end{enumerate}

\section{Optical Flow Based VSR}
Generally, most of the VSR methods contain three basic components (\textcolor{blue}{Lin et al. 2005}): (a) Motion compensation (alignment); (b) Interpolation; (c) Blur and noise removal (restoration).

\subsection{Motion Compensation}
The first component, i.e., motion compensation, which aims to obtain image alignment between multiple video frames, is a key component for VSR. Human vision system is sensitive to motion, thus capturing and modeling the influence of motion on visual perception is crucial for VSR (\textcolor{blue}{Li et al. 2016; Liu et al. 2018}). To get accurate video super-resolution result, the motion between video frames should be estimated as accurate as possible, because inaccurate motion distorts local structure and degrades the final HR image reconstruction (\textcolor{blue}{Su et al. 2012; Liao et al. 2015}). Consequently, how to perform a good motion estimation has attracted great attention.

Numerous motion estimation techniques, e.g., the global parametric models, and the local non-parametric models (e.g., block-motion approaches and optical flow based methods), have been proposed to improve the performance of video super-resolution (\textcolor{blue}{Schoenemann and Cremers 2012}).

Furthermore, sub-pixel shifts is another key factor for VSR (\textcolor{blue}{Babacan et al. 2011}). If the LR images only include integer shifts, there is no new information can be produced when combing them to reconstruct the HR image. In other word, capturing sub-pixel shifts is necessary in motion estimation for VSR (\textcolor{blue}{Dai et al. 2017}).

Since optical flow is able to supply accurate and sub-pixel motion information (\textcolor{blue}{Tu et al. 2014; Tu et al. 2019}), the optical flow based VSR method has been studied for a long time, leading to significant progresses in the past decade (\textcolor{blue}{Nguyen et al. 2018; Anwar et al. 2020}). Optical flow can model the temporal dependency between consecutive video frames (\textcolor{blue}{Wang et al. 2020b}), where the temporal dependency is normally considered as an essential component of VSR, and thus the estimation of optical flow can significantly affect the final result of VSR (\textcolor{blue}{Huang et al. 2018}). Besides, the optical flow field is a dense motion field that can describe the deformation or mapping of every pixel between two video frames, therefore the optical flow based VSR is very suitable for extracting the mapping of non-rigid moving objects in the video, and thus contributing to addressing the challenge of super-resolution for non-rigid objects in VSR.

\subsection{Optical Flow Based Motion Estimation and Compensation (MEC)}
\label{sec:5}

\subsubsection{Optical Flow (Motion) Estimation}
Optical flow estimation is based on the assumption that the brightness of a moving pixel remains constant over time. Mathematically, optical flow estimation is formulated as follows:
\begin{equation}
E(u,v)=F(I_t,I_{t+1};\Theta_F)
\label{OFE}
\end{equation}
where $I_t$ and $I_{t+1}$ are two successive input video frames, which separately denote the current frame at time $t$ (i.e. the target frame) and the next frame at time $t+1$ (i.e. the neighboring frame). $E$ refers to the estimation operation of optical flow, $F$ is a function utilized to calculate optical flow and $\Theta_F$ represents its parameters. $W = (u, v)$ denotes the calculated optical flow, with the horizontal and vertical flow components $u$ and $v$ respectively.

\begin{figure*}[!h]
 \centering
 \includegraphics[width=1.0\linewidth]{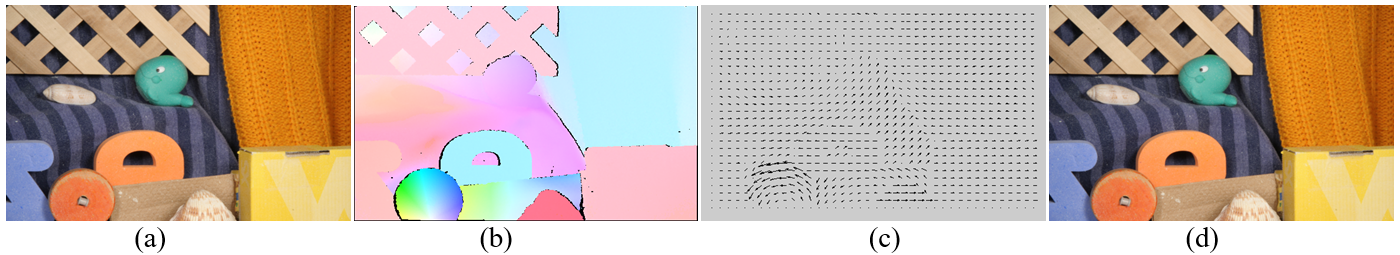}
 \caption{Optical flow: (a) and (d) respectively represents frame 10 and frame 11 of the RubberWhale sequence on the Middlebury benchmark (\textcolor{blue}{Baker et al. 2011}); (b) is the color-coded ground truth flow; (c) is the vector plot ground truth flow. The sample is taken from (\textcolor{blue}{Tu et al. 2019}).}
 \label{fig:OF-Illustration}
\end{figure*}

Figure~\ref{fig:OF-Illustration} shows the visualized optical flow. As (\textcolor{blue}{Tu et al. 2019}) stated: ``\textit{The 2D displacement field, which describes the apparent motion of brightness patterns between two successive images, is called the optical flow}." ``\textit{The optical flow field is ideally a dense field of displacement vectors (see Figure~\ref{fig:OF-Illustration} (b), (c)), which maps all points of the first image onto their corresponding locations in the second image}." Particularly, Figure~\ref{fig:OF-Illustration} (b) is the color-coded ground truth flow, and (\textcolor{blue}{Tu et al. 2019}) explained: ``\textit{The color-coded flow field is a dense visualization of the optical flow field. A color hue is associated to each direction and the saturation of the color increases with the magnitude of the flow vector}." Figure~\ref{fig:OF-Illustration} (c) is the vector plot ground truth flow, and (\textcolor{blue}{Tu et al. 2019}) described: ``\textit{which directly represents the displacement vectors and provides a good intuitive perception of physical motion}." To understand more knowledge about optical flow, please refer to the optical flow survey paper of (\textcolor{blue}{Tu et al. 2019}).

\vspace{0.5cm}
\subsubsection{Optical Flow Relied Motion Compensation}
\begin{figure*}[!htp]
 \centering
 \includegraphics[width=1.0\linewidth]{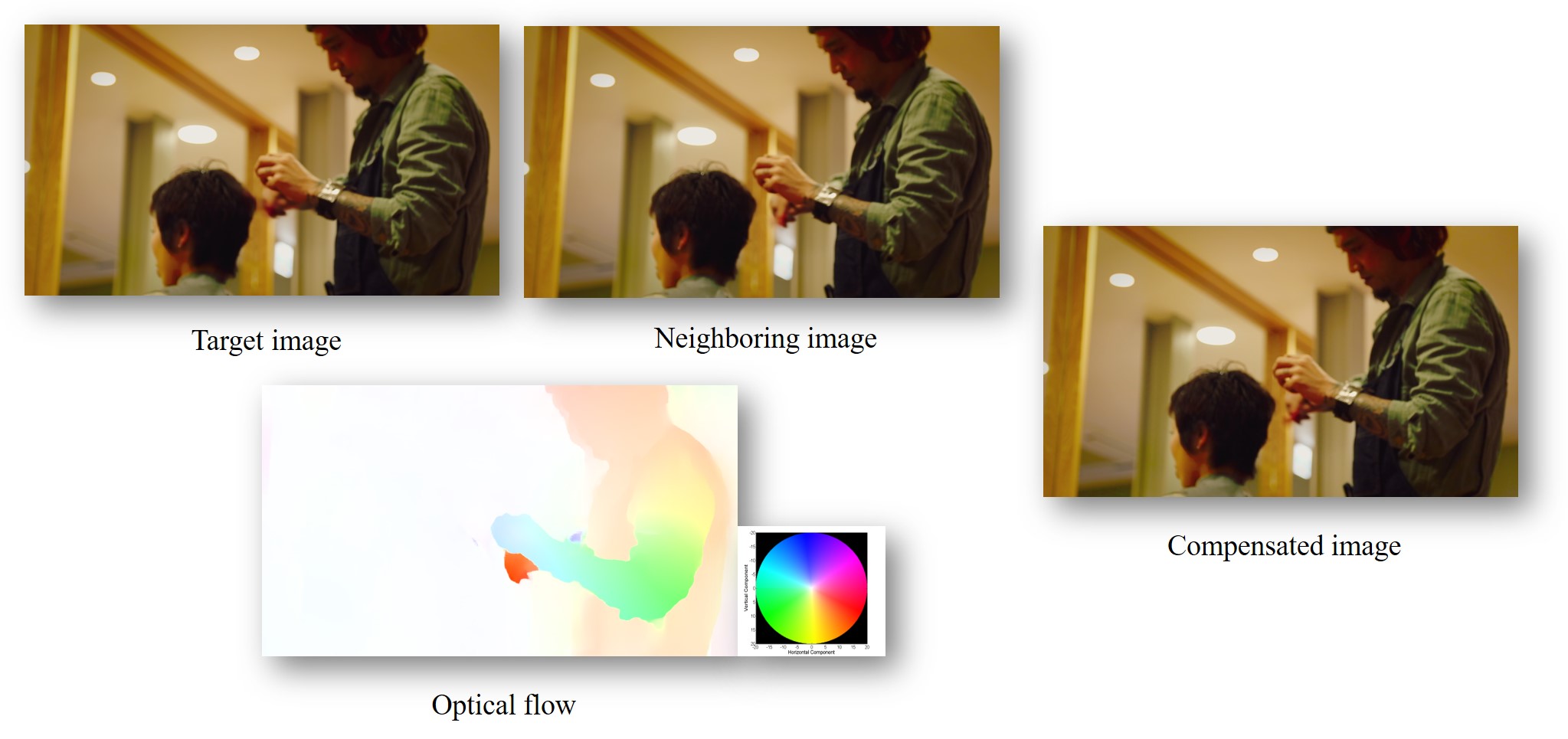}
 \caption{The general flowchart of motion estimation and compensation (MEC) in VSR.}
 \label{fig:MEC-VSR}
\end{figure*}

As shown in Figure~\ref{fig:MEC-VSR}, motion compensation is generally conducted by performing image transformation between an image sequence according to the estimated motion information to make neighboring frames matching with the target frame spatially. In particular, this operation can be achieved by certain methods, e.g., the geometric registration (\textcolor{blue}{Nasrollahi and Moeslund 2014; Liu et al. 2020}). Mathematically, a compensated frame is represented as:
\begin{equation}
I_{MC}=F_{MC}(I,W;\theta_{ME})
\label{CompensatedF}
\end{equation}
where $F_{MC}(\cdot)$ denotes a motion compensation function, $I$ is the neighboring frame, $W$ represents the estimated optical flow, and $\theta_{ME}$ denotes the parameters of optical flow based motion estimation. Please refer to Figure~\ref{fig:MEC-VSR} for motion estimation and motion compensation in detail.

\textit{\textbf{Summary}}. Motion cue plays a crucial role in capturing temporal dependency between LR video frames for VSR:
\begin{itemize}
\item Motion compensation, which is one of the main components in VSR, encodes temporal dependency in compensated LR video frames in terms of estimating temporal information from consecutive frames;
\item Optical flow, which supplies accurate, sub-pixel and dense motion information, as well as also can capture the motion of objects that are non-rigid, non-planar and self-occluded (\textcolor{blue}{Nasrollahi and Moeslund 2014}), is good for modeling the temporal information.
\end{itemize}

\section{Temporal Dependency Capturing for VSR with OF}
\label{sec:6}
\begin{figure*}[!h]
 \centering
 \includegraphics[width=1.0\linewidth]{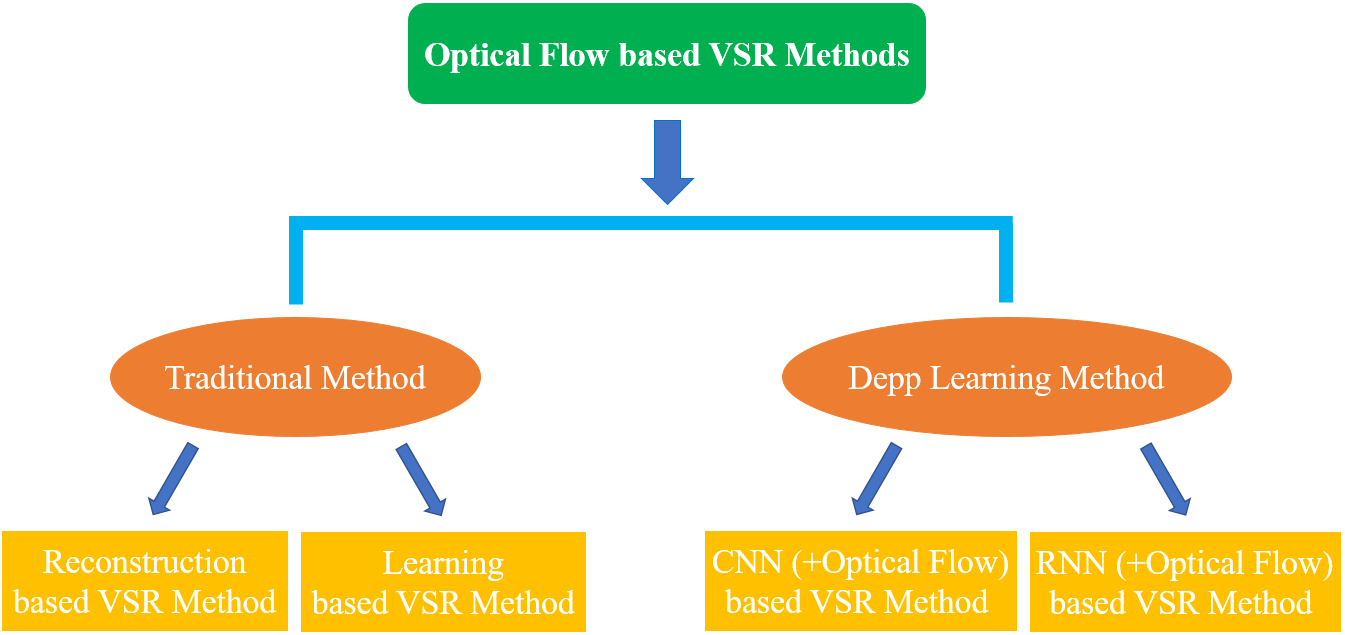}
 \caption{The categories of optical flow based VSR methods.}
 \label{fig:VSR-Category}
\end{figure*}

In contrast to a single image, adjacent video frames provide temporal correlations. Therefore, to conduct VSR, it is essential to exploit temporal dependency between consecutive frames efficiently and effectively (\textcolor{blue}{Wang et al. 2020b}). To model the temporal dependency, optical flow is extensively utilized (\textcolor{blue}{Caballero et al. 2017; Liu et al. 2018; Wang et al. 2020b}). The VSR methods, which used optical flow to capture the temporal dependency, can be broadly classified into three main categories: (1) reconstruction based VSR method, (2) learning-based VSR method and, (3) deep learning based VSR method. While the first two approaches can be treated as the traditional VSR method, the deep learning based VSR method can be further classified into two groups: (a) the CNN (+optical flow) based VSR method and; (b) the RNN (+optical flow) based VSR method. Figure~\ref{fig:VSR-Category} shows the categories of optical flow based VSR methods. The first three VSR methods always have explicit motion compensation, while the RNN based VSR methods normally do not have explicit motion compensation. The first two traditional VSR methods have been studied for decades, but are surpassed by their deep learning based counterparts.

\textbf{\textit{Summary}}. Most of the optical flow related VSR methods exploit motion information in two main aspects:
\begin{itemize}
\item Explicitly registering multiple video frames in terms of the estimated motion;
\item Implicitly embedding motion estimation to regularize the process of HR image reconstruction.
\end{itemize} 

\subsection{Reconstruction based VSR Method}
\label{sec:7}
To model the temporal dependency in VSR, some algorithms first perform optical flow estimation explicitly to compute sub-pixel motion between consecutive LR video frames and then warp each LR image to the target HR space according to the computed optical flow. In this way, the difference caused by the movement between the LR video frames can be captured, and the correspondence between the LR observations and the desired reconstructed HR image can be exploited, where they are useful for guiding the reconstruction of the targeted HR image. 
This kind of algorithms is reconstruction ($+$ optical flow) based VSR methods, which will be called reconstruction-based VSR methods for short (\textcolor{blue}{Mitzel et al. 2009; Huang et al. 2015; Liao et al. 2015}).

\begin{figure}
 \centering
 \includegraphics[width=1.0\linewidth]{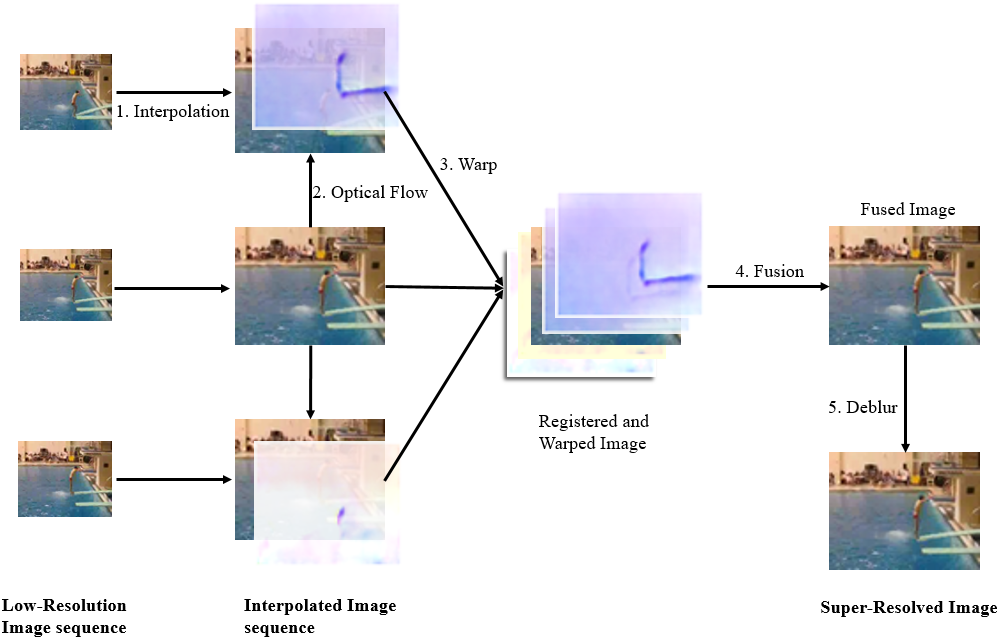}
 \caption{The flow-chat of the Reconstruction (+ optical flow) based VSR method. This method usually includes 5 steps: 1) Performing interpolation to the LR image sequence to get higher resolution image sequence; 2) Computing optical flow between successive images; 3) Registering images to the reference image by using the motion information of the estimated optical flow; 4) Estimating the SR image by fusing the reference image and the registered images; 5) Recovering the final SR image via deblurring.}
 \label{fig:ReconSR}
\end{figure}

For the reconstruction-based VSR methods involving optical flow estimation, since LR video frames have different sub-pixel motions and rotations from each other, it is crucial to obtain motion information precisely before fusing them to produce an HR image. Inaccurate motion cue will lead to various types of visual artifacts that subsequently damage the quality of the reconstructed HR image (\textcolor{blue}{Thapa et al. 2016}). In brief, the accuracy and efficiency of optical flow estimation critically affects the performance of the reconstruction-based VSR. Figure~\ref{fig:ReconSR} shows the framework of the Reconstruction (+ optical flow) based VSR method.

\textcolor{blue}{Baker and Kanade 1999} proposed a pioneer work to estimate optical flow for VSR to address the issue of complex motions in realistic videos. The correlation between super-resolution optical flow and pyramid-based image representations is analyzed. \textcolor{blue}{Zhao and Sawhney 2002} gave a systematic study to discover the impact of image alignment and warping errors on the VSR. Specifically, the performance of VSR under alignment with piecewise parametric or optical flow based approaches are investigated. They revealed that optical flow is especially significant for reconstructing high-frequency components in the HR image, and the flow consistency and flow accuracy are two critical elements. Optical flow consistency across video frames are important, and optical flow errors could cause super-resolution to become infeasible. \textcolor{blue}{Mitzel et al. 2009} proposed a variational framework for VSR with arbitrary videos. This work contains two important steps: firstly, a quadratic relaxation strategy, which is able to compute high accuracy optical flow, is introduced to get motion for mapping LR images. Secondly, a variational method, which can impose a total variation regularity to the computed intensity map, is used to estimate the HR image. \textcolor{blue}{Keller et al. 2011} exploited a full motion-compensated variational framework to design a simultaneous VSR method, which is able to jointly compute the HR image sequence and its corresponding HR optical flow. In this way, the VSR performance is boosted as more accurate details can be propagated from frame to frame. There are two main contributions of \textcolor{blue}{Keller et al. 2011}: (1) it is the first work to calculate super-resolved optical flows; (2) this method is possible to be used to general video with arbitrary scene content and/or arbitrary optical flows. In particular, the optical flow is estimated according to a classical total variational energy function (\textcolor{blue}{Papenberg et al. 2006}).

On the other side, to alleviate the degradation caused by the estimated inaccurate dense optical flow in the reconstruction based VSR method, \textcolor{blue}{Su et al. 2012} proposed to compute local flow with reliable accuracy based on the sparse feature point (e.g. SIFT, corner point) correspondences. This strategy is effective even when the input video frames are with noise, large-scale or other complex local motion.

\textbf{\textit{Limitation}}. Reconstruction-based VSR methods with explicit optical flow estimation generally suffer from the following limitations:

\begin{itemize}
   \item High computational cost for optical flow estimation;
   \item Difficulty in obtaining high quality motion information even with the state-of-the-art optical flow approaches;
   \item Visual artifacts inevitably caused by inaccurate registration of erroneous motions during the reconstruction;
   \item Degenerated cases: Inaccurate optical flow may sometimes lead a poor reconstruction performance worse than direct image interpolation.
\end{itemize}

\subsection{Learning based VSR method}
\label{sec:8}
Reconstructing the HR image from LR video frames is an ill-posed problem and needs to be regularized (\textcolor{blue}{Kappeler et al. 2016}). Consequently, some probabilistic models, e.g., Expectation-Maximization (EM) framework and Bayesian framework, are proposed to introduce priors to control the smoothness or the total variation of the image (\textcolor{blue}{Mudenagudi et al. 2011}). This kind of VSR approaches are considered as the learning (+ optical flow) based VSR method, called learning based VSR method for short (\textcolor{blue}{Liu and Sun 2014; Ma et al. 2015}).

Generally, VSR methods make some simplifying assumptions. For example, the underlying motion may be assumed to have an oversimplified parametric form, or the blur kernel and noise levels are assumed to be known. But in practice, the movement of objects and cameras can be arbitrary, the motion blur and point spread functions can result in an unknown blur kernel, and the noise levels in video are unknown. Therefore, the learning based VSR methods attempt to integrate all these factors in a single framework without making oversimplified assumptions, and optimized them simultaneously.

\textcolor{blue}{Liu and Sun 2014} exploited a Bayesian method for adaptive VSR by estimating optical flow, blur kernel and noise level in addition to reconstruct the original HR video frames simultaneously in a single framework. They optimize the optical flow and the noise level jointly in a coarse-to-fine manner on a Gaussian image pyramid. At each pyramid level, the optical flow and noise level are computed iteratively in the maximum a posterior (MAP) way. To address the issue of motion blur in VSR, \textcolor{blue}{Ma et al. 2015} proposed to search least blurred pixels in VSR optimally. An EM framework was designed to guide residual blur estimation and HR image reconstruction. The classical optical flow regularizer (\textcolor{blue}{Sun et al. 2010}) was selected for supplying the motion prior. To reduce the computational cost of motion estimation, they used the interpolated TV-L1 optical flow (\textcolor{blue}{Sun et al. 2010}) on the LR images for approximation.

\textbf{\textit{Limitation}}. The learning-based VSR methods usually formulate VSR as an optimization problem, and estimate the HR image, optical flow and blur kernel alternately or simultaneously. Since a large number of iterations are needed to reach convergence, the learning-based VSR methods are also time-consuming.

\subsection{Deep Learning based VSR Method}
\label{sec:9}
As we analyzed above, the hand-crafted VSR approaches treat SR as a sophisticated optimization problem, which require expensive computational time and suffer considerable inference cost. Furthermore, the hand-crafted VSR approaches are not always applicable for practical scenarios where the imaging process may have different properties than assumed in the learning stage, leading to degraded performance (\textcolor{blue}{Yang et al. 2018}).
\begin{figure}
 \centering
 \includegraphics[width=1.0\linewidth]{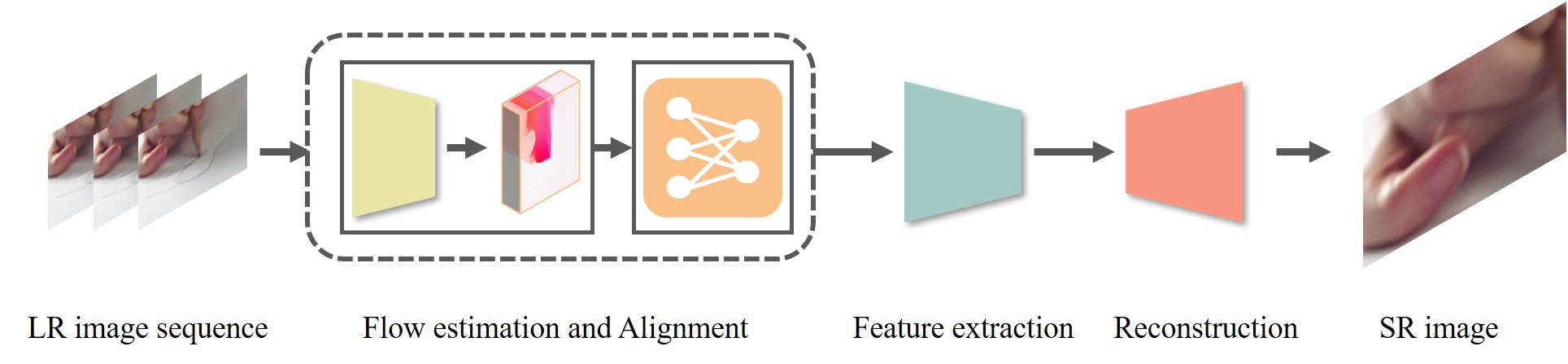}
 \caption{The flow-chat of the Deep Learning (+ optical flow) based VSR method. Specifically, in the ``Flow estimation and Alignment" module, optical flow estimation is conducted on LR video frames for motion compensation, and the alignment is achieved by learning a mapping from compensated LR video frames. In the ``Feature extraction" module, the deep feature is extracted with a deep neural network (e.g. CNN, RNN). In the ``Reconstruction" module, the HR video frames are super-resolved for the corresponding LR video frames.}
 \label{fig:DeepLOF-SR}
\end{figure}

Recently, deep learning based VSR methods have been proposed, with explicit or implicit temporal alignment. They have become the dominant technique for VSR (\textcolor{blue}{Jo et al. 2018; Lucas et al. 2019}) since deep networks have strong model capacity to learn useful priors for VSR from a large video dataset, and can be trained end-to-end. Figure~\ref{fig:DeepLOF-SR} shows the framework of the Deep Learning (+ optical flow) based VSR method.

According to how the temporal dependency among successive LR frames is exploited, the deep learning based VSR techniques have two main strategies (\textcolor{blue}{Liao et al. 2015; Kappeler et al. 2016; Liu et al. 2018}): (a) utilizing convolutional neural networks (CNNs), and performing motion compensation explicitly to align LR video frames as the input for the CNN model (\textcolor{blue}{Liu et al. 2018}), (b) using recurrent networks (RNNs) to capture the temporal dependency (\textcolor{blue}{Haris et al. 2019}) to avoid perform motion estimation explicitly.

\subsubsection{CNN ($+$ Optical Flow) based VSR Method}
\label{sec:10}
The CNN based VSR methods are usually conducted in the following way: firstly, estimating optical flow from LR video frames for motion compensation; secondly, adjacent LR frames are motion compensated and utilized as the input to a CNN to produce HR images (\textcolor{blue}{Kappeler et al. 2016; Sajjadi et al. 2018; Wang et al. 2020b}). We termed this kind of approaches as the CNN (+ Optical Flow) based VSR method, called CNN based VSR method for short.

Since \textcolor{blue}{Dong et al. 2014} proposed to use CNN to conduct single image SR and achieved the state-of-the-art performance, CNN is widely investigated for SR. Currently, the CNN based VSR method has become the dominant technique due to the following advantages:
\begin{itemize}
\item Once a CNN is trained, super-resolving an image is just a feed-forward process, making it is much more efficient than the traditional VSR methods;
\item Neural networks have powerful learning capacity to model the spatial relation of the video frames, especially when with sufficient video data;
\item The CNN framework usually can be trained end-to-end.
\end{itemize}

~\\
\noindent \textbf{(A). Temporal Concatenation}

~\\
To model temporal information in VSR, one of the most popular methods is to concatenate the frames (\textcolor{blue}{Liao et al. 2015; Kappeler et al. 2016; Caballero et al. 2017; Wang et al. 2020b}). However, this strategy is unable to represent multiple motion regimes on a sequence as the input video frames are directly concatenated together (\textcolor{blue}{Haris et al. 2019}). Furthermore, it is hard to train the network since many frames are processed simultaneously.

\begin{figure}[!h]
 \centering
 \includegraphics[width=0.85\linewidth]{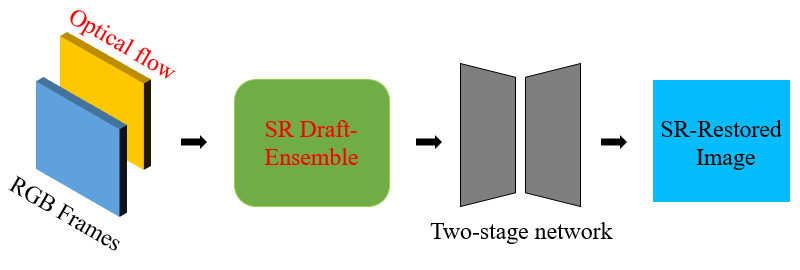}
 \caption{The architecture of the Deep-DE model.}
 \label{fig:Deep-DE}
\end{figure}

\textbf{\textit{Deep-DE}}: \textcolor{blue}{Liao et al. 2015} proposed a deep draft-ensemble (Deep-DE) learning SR framework for fast VSR. They integrate SR drafts via the nonlinear process in a convolutional neural network (CNN) to recover high-frequency details. The SR draft-ensemble process can generate several SR drafts according to a set of motion estimates, i.e., optical flows, where the optical flows are computed via different parameter settings. The architecture of the Deep-DE model is depicted in Figure~\ref{fig:Deep-DE}.

\begin{figure}[!h]
 \centering
 \includegraphics[width=0.75\linewidth]{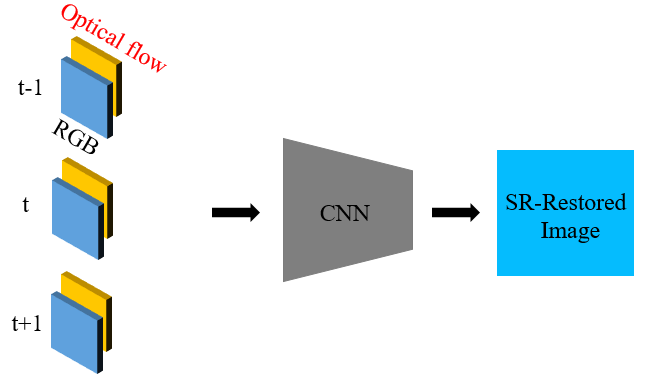}
 \caption{The architecture of the VSR-Net model.}
 \label{fig:VSR-Net}
\end{figure}

\textbf{\textit{VSR-Net}}: \textcolor{blue}{Kappeler et al. 2016} presented to use CNN for VSR, where the CNN is trained on both the spatial and the temporal dimensions of videos to boost their spatial resolution. To obtain motion compensated frames as input for CNN, an adaptive motion compensation approach is introduced, in which the motions of input LR frames are compensated via a traditional optical flow algorithm. Then, the compensated frames are concatenated and fed to a pre-trained CNN SR network to reconstruct the SR frame. The optical flow algorithm (\textcolor{blue}{Drulea and Nedevschi 2011}), which is a combination of the Local-Global approach with Total Variation (CLG-TV), is used. Besides, this adaptive motion compensation approach is able to address the issues of fast moving objects and motion blur in videos. The architecture of the VSR-Net model is depicted in Figure~\ref{fig:VSR-Net}.

\begin{figure}[!h]
 \centering
 \includegraphics[width=0.75\linewidth]{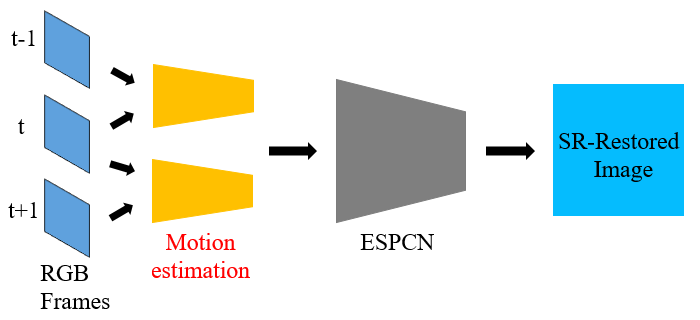}
 \caption{The architecture of the VESPCN model.}
 \label{fig:VESPCN}
\end{figure}

\textbf{\textit{VESPCN}}: Methods Deep-DE (\textcolor{blue}{Liao et al. 2015}) and VSR-Net (\textcolor{blue}{Kappeler et al. 2016}) use a two-step framework, which separate the motion estimation from the network training. Consequently, they are hard to get an overall optimal solution. To address this issue, \textcolor{blue}{Caballero et al. 2017} proposed a VESPCN method, which exploits a trainable motion compensation network, and utilizes a CNN to produce HR predictions from multiple LR frames in an end-to-end manner. It is the first end-to-end CNN based VSR framework. It takes advantage of sub-pixel convolution, temporal redundancy extraction via the spatio-temporal network, and the motion compensation, resulting in improved VSR performance in both accuracy and efficiency. Particularly, they designed an efficient spatial motion compensation transformer (MCT) module to estimate and compensate the motion between video frames in terms of optical flow, where the optical flow is computed in a coarse-to-fine strategy. After that the compensated frames are fed into the convolutional network for feature extraction and fusion. At last, the super-resolution process is conducted through a sub-pixel convolutional layer. The architecture of the VESPCN model is depicted in Figure~\ref{fig:VESPCN}.

\begin{figure}[!h]
 \centering
 \includegraphics[width=0.75\linewidth]{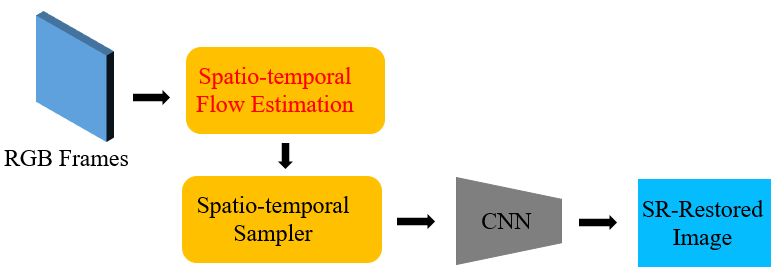}
 \caption{The architecture of the STTN model.}
 \label{fig:STTN}
\end{figure}

\textbf{\textit{STTN}}: Most optical flow techniques process only a pair of video frames (\textcolor{blue}{Kim et al. 2018}), which are sensitive to complex situations like noise, occlusion and illumination change, and thus optical flow is limited for VSR. To overcome this disadvantage of the optical flow approaches, \textcolor{blue}{Kim et al. 2018} presented a spatio-temporal transformer network (STTN), which is able to handle multiple frames at a time. STTN consists of three main components: (a) a spatiotemporal flow estimation module, (b) a spatio-temporal sampler module, and (c) a super-resolution module. In particular, (a) the spatio-temporal flow estimation module is a U-Net style network (Ronneberger et al. 2015), which can estimate the optical flow of successive input frames including the target frame and multiple neighboring frames. The final result is a 3-channel spatio-temporal flow that describes the spatial and temporal changes between multiple video frames. (b) The spatio-temporal sampler module is in fact a trilinear interpolation approach, which is used to conduct warping for the current multiple neighboring frames and obtain the aligned video frames in terms of the spatio-temporal optical flow that is gained by the spatio-temporal flow module. (c) The super-resolution module is applied to perform feature fusion and super-resolution reconstruction for the target frame. The architecture of the STTN model is depicted in Figure~\ref{fig:STTN}.

\begin{figure}[!h]
 \centering
 \includegraphics[width=0.75\linewidth]{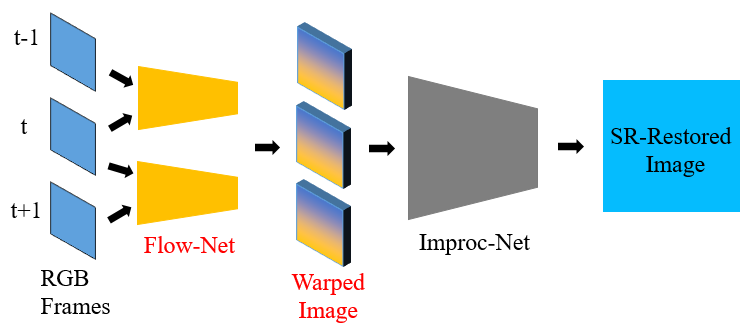}
 \caption{The architecture of the TOFlow model.}
 \label{fig:TOFlow}
\end{figure}

\textbf{\textit{TOFlow}}: \textcolor{blue}{Xue et al. 2019} designed a task-oriented flow (TOFlow) architecture, which combines the optical flow estimation network with the super-resolution reconstruction network, and trains these two networks jointly to compute optical flow. TOFlow adopts the framework of SpyNet (\textcolor{blue}{Ranjan and Black 2017}) for optical flow estimation, and uses a spatial transformer approach to warp the neighboring frames based on the calculated optical flow. The final super-resolution is implemented by a image processing module. The architecture of the TOFlow model is depicted in Figure~\ref{fig:TOFlow}.

\begin{figure}[!h]
 \centering
 \includegraphics[width=0.75\linewidth]{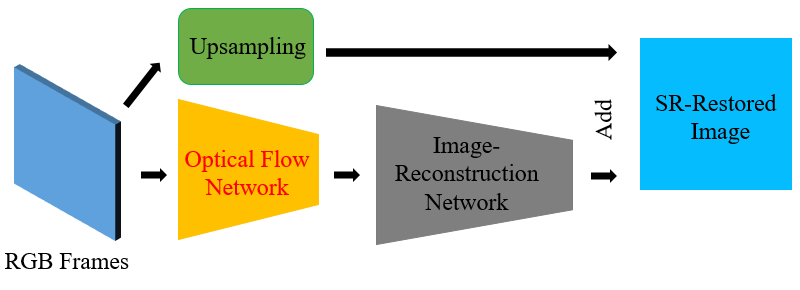}
 \caption{The architecture of the MMCNN model.}
 \label{fig:MMCNN}
\end{figure}

\textbf{\textit{MMCNN}}: To better extract spatiotemporal correlations between successive LR video frames and find more realistic details, \textcolor{blue}{Wang et al. 2019} presented a multi-memory CNN (MMCNN) for VSR, in which an optical flow network and an image-reconstruction network are cascaded. By embedding convolutional long short-term memory into the residual block, they designed a multi-memory residual block to replace the ordinary single-memory module to learn and retain inter-frame temporal correlations between the adjacent LR frames gradually. Specifically, the optical flow network can allow consecutive frames to serve as reference frames, thus it is beneficial for fusing multi-frame information. The architecture of the MMCNN model is depicted in Figure~\ref{fig:MMCNN}.

\begin{figure}[!h]
 \centering
 \includegraphics[width=0.75\linewidth]{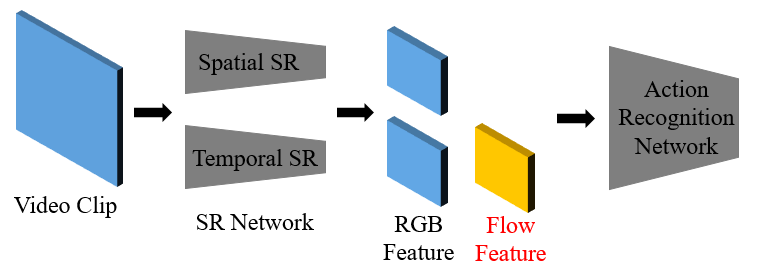}
 \caption{The architecture of the SoSR-ToSR model.}
 \label{fig:SoSR-ToSR}
\end{figure}

\textbf{\textit{SoSR-ToSR}}: \textcolor{blue}{Zhang et al. 2019} applied VSR as a preprocessing step before feeding LR video frames into a two-stream action recognition network to address the issue that action recognition methods are un-applicable on low resolution videos. Specifically, to improve the performance of VSR, for the spatial stream, they designed an optical flow guided weighted MSE loss to guide a spatial-oriented SR (SoSR) network to focus more on the regions with motion. For the temporal stream, they exploited a temporal-oriented SR (ToSR) network to enhance the adjacent frames together to ensure the temporal consistency. The architecture of the SoSR-ToSR model is depicted in Figure~\ref{fig:SoSR-ToSR}.

\begin{figure}[!h]
 \centering
 \includegraphics[width=0.75\linewidth]{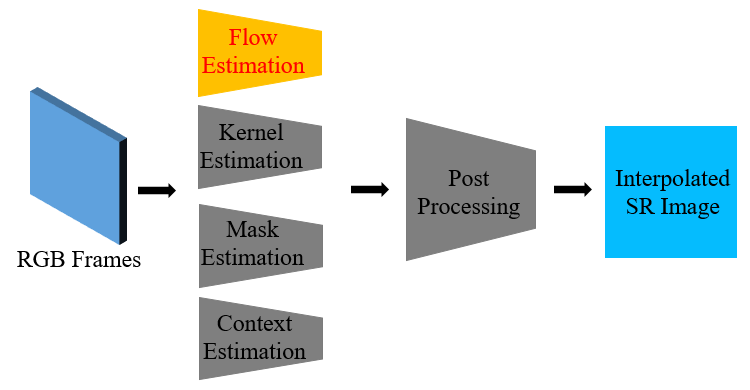}
 \caption{The architecture of the MEMC-Net model.}
 \label{fig:MEMC-Net}
\end{figure}

\textbf{\textit{MEMC-Net}}: Inspired by EDSR (\textcolor{blue}{Lim et al. 2017}), \textcolor{blue}{Bao et al. 2019} proposed a motion estimation and motion compensation network (MEMC-Net) for VSR. The main contribution of MEMC-Net is the exploited adaptive warping layer, which warps the neighboring video frames by the computed optical flow and the convolutional kernel. The FlowNet (\textcolor{blue}{Dosovitskiy et al. 2015}) is used for the motion estimation network, and a modified U-Net (\textcolor{blue}{Ronneberger et al. 2015}), which consists of five max-pooling layers, five un-pooling layers and skip connections from the encoder to the decoder, is utilized for the kernel estimation network. To handle the occlusion problem, MEMC-Net extracts the feature of input frames by a pre-trained ResNet18, and feeds the output of the first convolutional layer of ResNet18 as the context information into the adaptive warping layer to conduct the warping again. the architecture of the MEMC-Net model is depicted in Figure~\ref{fig:MEMC-Net}.

\begin{figure}[!h]
 \centering
 \includegraphics[width=0.75\linewidth]{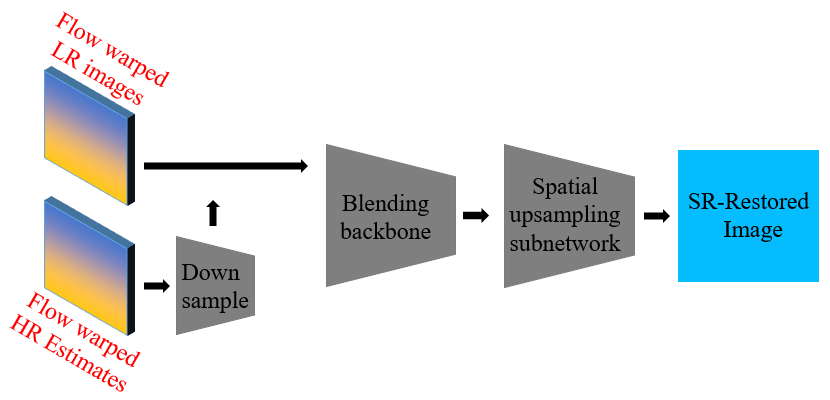}
 \caption{The architecture of the MultiBoot-VSR model.}
 \label{fig:MultiBoot-VSR}
\end{figure}

\textbf{\textit{MultiBoot-VSR}}: \textcolor{blue}{Kalarot et al. 2019} proposed a multi-stage multi-reference bootstrapping method for VSR (MultiBoot-VSR). MultiBoot-VSR is a two-stage framework, where the output of the first stage is utilized as the input of the second stage to further boost the performance. Initially, the FlowNet 2.0 (\textcolor{blue}{Ilg et al. 2017}) algorithm is adopted to estimate optical flow and operate motion compensation. After that, the processed video frames are fed into the first-stage network to super-resolve the target frame. Lastly, the output from the first-stage is downsampled, concatenated with the original LR frame, and input to the second stage MultiBoot network to compute the final super-resolution result of the target frame. The architecture of the MultiBoot-VSR model is depicted in Figure~\ref{fig:MultiBoot-VSR}.

\begin{figure}[!h]
 \centering
 \includegraphics[width=0.75\linewidth]{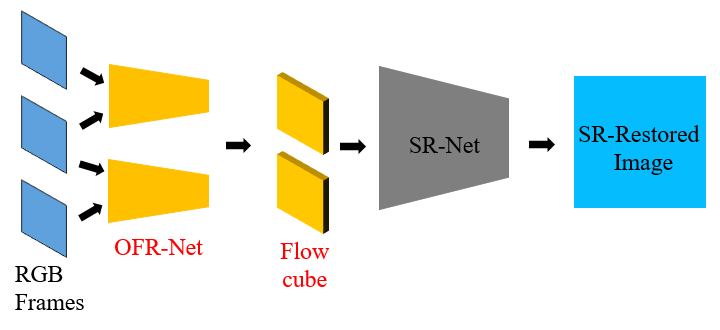}
 \caption{The architecture of the SOF-VSR model.}
 \label{fig:SOF-VSR}
\end{figure}

\textbf{\textit{SOF-VSR}}: To handle the problem of resolution conflict between LR optical flows and HR outputs, where the resolution conflict prevents the recovery of fine temporal details, \textcolor{blue}{Wang et al. 2020b} presented an end-to-end network to super-resolve optical flows for VSR, which is named SOF-VSR. In particular, this work contains three main steps: (a) an optical flow reconstruction network (OFR-Net) is used to infer HR optical flows in a coarse-to-fine way; (b) motion compensation is conducted via the estimated HR optical flows to encode temporal dependency; (c) the compensated LR images are input to a super-resolution network (SR-Net) to produce SR images. The architecture of the SOF-VSR model is depicted in Figure~\ref{fig:SOF-VSR}.

\begin{figure}[!h]
 \centering
 \includegraphics[width=0.75\linewidth]{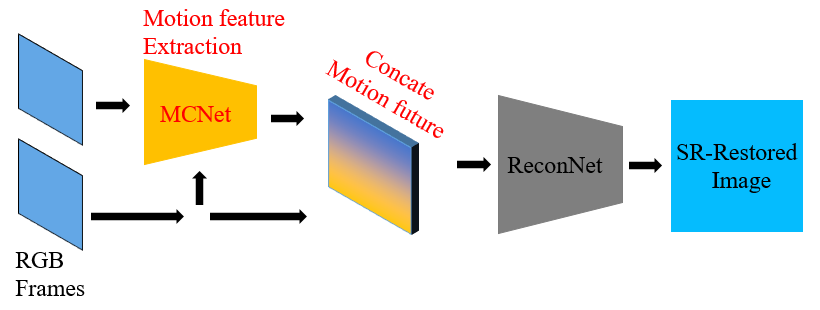}
 \caption{The architecture of the DDAN model.}
 \label{fig:DDAN}
\end{figure}

\textbf{\textit{DDAN}}: \textcolor{blue}{Li et al. 2020} explored a deep dual attention network (DDAN), which consists of two main components, i.e., a motion compensation network (MCNet) and a SR reconstruction network (ReconNet), to fully exploit the spatio-temporal dependencies and learn discriminative spatio-temporal features for accurate VSR. To be specific, (1) the MCNet progressively learns multi-scale optical flow representations to synthesize the motion information across adjacent frames in a pyramid coarse-to-fine manner. To reduce the mis-registration errors caused by the optical flow based motion compensation, DDAN (\textcolor{blue}{Li et al. 2020}) took two effective measures. First, except adopting the commonly used downscaling motion estimation strategy, it also utilizes a module without any downsampling operation to capture full resolution optical flow representations for better motion compensation. Second, the prior optical flow based methods usually simply concatenate the compensated frames and center frame for feature extraction and reconstruction, where the errors in the estimated optical flow or wrapping will adversely impact the subsequent SR reconstruction and bring artifacts. In contrast, DDAN extracts the detail components of original LR neighboring frames as complementary information to alleviate the errors of motion estimation. (2) In the ReconNet, DDAN incorporates the dual attention mechanism along channel and spatial dimensions with residual learning to focus on the intermediate informative features for high-frequency details recovery. The MCNet and ReconNet can be trained jointly in an end-to-end way for motion compensation and video SR reconstruction. The architecture of the DDAN model is depicted in Figure~\ref{fig:DDAN}.

~\\
~\\
\noindent \textbf{(B). Temporal Aggregation}

~\\
To address the issue of dynamic motion in VSR, some methods presented multiple SR inferences to work on different motion regimes (\textcolor{blue}{Liu et al. 2018}), where the outputs of all branches are aggregated at the last layer to construct the SR frame. However, these methods are difficult for global optimization because they also need to concatenate some input frames.

\begin{figure}[!h]
 \centering
 \includegraphics[width=0.75\linewidth]{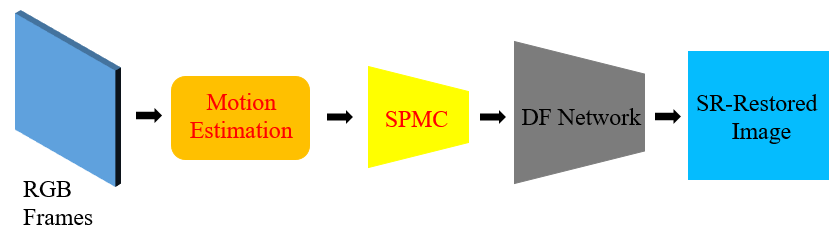}
 \caption{The architecture of the DRVSR model.}
 \label{fig:DRVSR}
\end{figure}

\textbf{\textit{DRVSR}}: To address two important sub-problems in VSR, i.e., (a) aligning multiple frames to construct accurate correspondence and (b) fusing image details to produce high-quality results, \textcolor{blue}{Tao et al. 2017} proposed a detail-revealing deep video super-resolution (DRVSR) method, in which a ``sub-pixel motion compensation" (SPMC) layer is exploited to conduct the up-sampling and motion compensation jointly for adjacent input frames according to the estimated optical flow. Specifically, DRVSR includes three main components, i.e., (a) a motion estimation module, (b) a motion compensation module, and (c) a fusion module. DRVSR respectively uses the motion compensation transformer (MCT) (\textcolor{blue}{Caballero et al. 2017}) for motion estimation and the SPMC layer for motion compensation. Specifically, the SPMC layer applies sub-pixel information from the optical flow field to get sub-pixel motion compensation and resolution enhancement. For the fusion module, a detail fusion (DF) network is utilized to fuse image details from multiple video frames after SPMC alignment effectively. Additionally, the fusion module also utilizes a ConvLSTM module (\textcolor{blue}{Glorot and Bengio 2010}) to tackle the spatio-temporal information. However, the SPMC layer costs large memory and has limited function, which prevents DRVSR to go deeper. The architecture of the DRVSR model is depicted in Figure~\ref{fig:DRVSR}.

\begin{figure}[!h]
 \centering
 \includegraphics[width=0.75\linewidth]{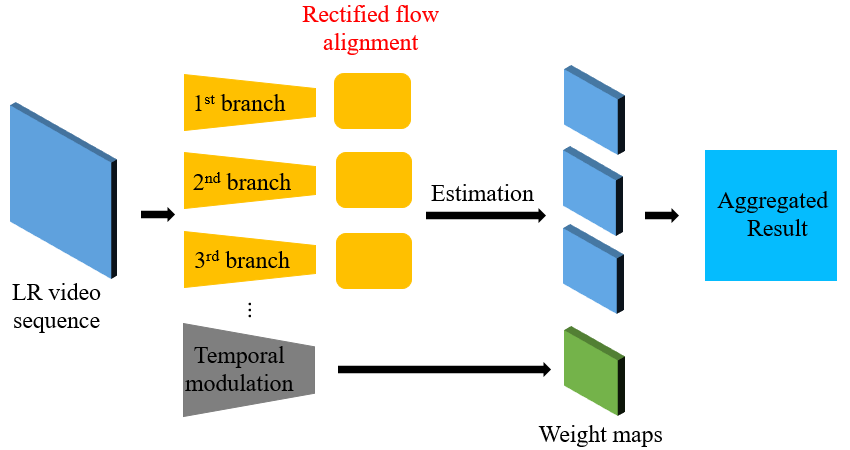}
 \caption{The architecture of the TANN model.}
 \label{fig:TANN}
\end{figure}

\textbf{\textit{TANN}}: \textcolor{blue}{Liu et al. 2017} presented a temporal adaptive neural network (TANN), which can tackle various types of movement robustly and choose the optimal range of temporal dependency automatically. In this way, useful information among successive video frames can be captured and the damage caused by erroneous motion can be reduced. They simplify the motion estimation in the patch level to incorporate translations to avoid interpolation. The rectified optical flow alignment is better than the traditional optical flow based image alignment in reconstructing the HR image. The architecture of the TANN model is depicted in Figure~\ref{fig:TANN}.

\textbf{\textit{Limitation}}. The CNN based VSR methods generally suffer from the following limitations:
\begin{itemize}
\item Dividing the VSR into a large number of separate multi-frame SR subtasks, resulting in temporal inconsistent, and unsatisfactory flickering artifacts would be produced.

\item For the commonly used frame-concatenation strategy, many frames are processed simultaneously in the network, leading to hard training.

\item Much time is wasted as each input frame is processed several times in the network.
\end{itemize}

\subsubsection{RNN (+ Optical Flow) based VSR Method}
\label{sec:11}
Optical flow estimation 
usually costs high computational cost. To boost the efficiency, recurrent neural networks (RNNs) are now widely used for VSR, after the pioneering work of (\textcolor{blue}{Huang et al. 2015; Xiang et al. 2020}). Generally, compared to CNN based VSR approaches, in the RNN based VSR framework, implicit temporal alignment is performed in terms of optical flow to replace explicit temporal alignment that depends on optical flow based motion estimation and compensation (\textcolor{blue}{Sajjadi et al. 2018; Yang et al. 2018}). We termed this kind of approaches as RNN (+ Optical Flow) based VSR method, called RNN based VSR method for short.

\begin{figure}[!h]
 \centering
 \includegraphics[width=0.75\linewidth]{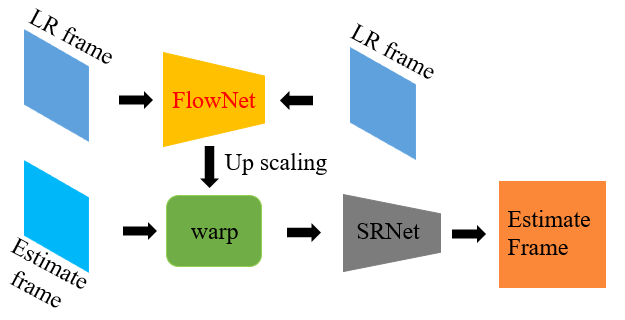}
 \caption{The architecture of the FRVSR model.}
 \label{fig:FRVSR}
\end{figure}

\textbf{\textit{FRVSR}}: \textcolor{blue}{Sajjadi et al. 2018} exploited an effective end-to-end trainable frame-recurrent VSR method (named FRVSR), which uses previously inferred HR estimates to super-resolve the subsequent video frames. Due to the application of the recurrent architecture, two benefits are gained: (a) Reducing the computational cost as each input frame is only processed once. (b) Enhancing the ability of the network to produce temporally consistent frames. Because the information from past frames will be passed to later frames through the HR estimates which are recurrently propagated over time. The FRVSR framework contains two important components, i.e., (a) the optical flow estimation network FlowNet, and (b) the super-resolution network SRNet. One distinct characteristic of FRVSR is its alignment strategy, where it does not warp the prior frame of the target directly while warps the HR version of the prior frame instead. However, since FRVSR simply refers to previously inferred HR frames, serious jitter and jagged artifacts are generated due to the former super-resolving errors are accumulated to the subsequent frames. The architecture of the FRVSR model is depicted in Figure~\ref{fig:FRVSR}.

\begin{figure}[!h]
 \centering
 \includegraphics[width=0.75\linewidth]{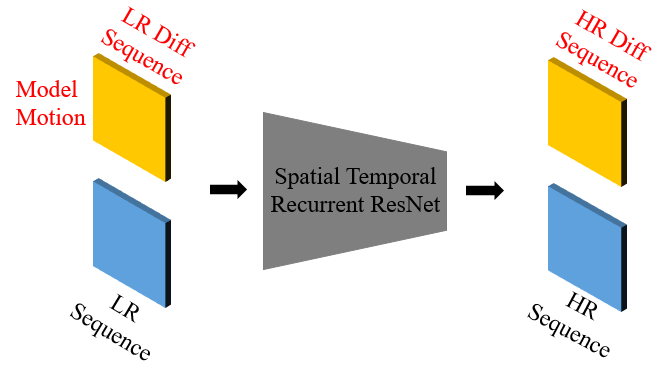}
 \caption{The architecture of the STR-ResNet model.}
 \label{fig:STR-ResNet}
\end{figure}

\textbf{\textit{STR-ResNet}}: \textcolor{blue}{Yang et al. 2018} exploited a Spatial Temporal Recurrent Residual Network (STR-ResNet) for video SR, which is able to model intra-frame redundancy and inter-frame motion context jointly in a unified deep framework, due to the framework combines the spatial convolutional and temporal recurrent architectures. This network does not require explicit optical flow estimation for motion compensation. The architecture of the STR-ResNet model is depicted in Figure~\ref{fig:STR-ResNet}.

\begin{figure}[!h]
 \centering
 \includegraphics[width=0.75\linewidth]{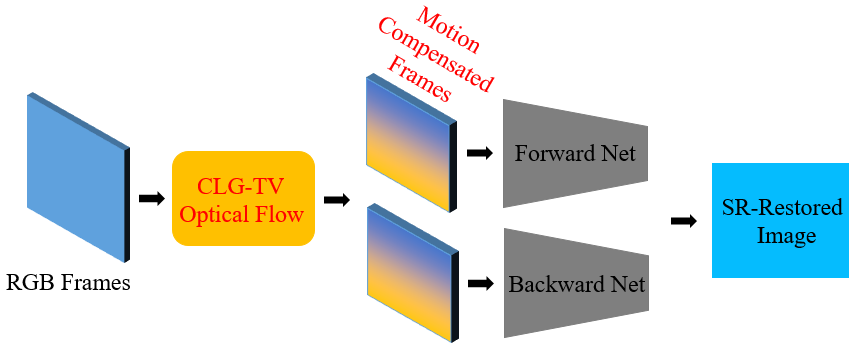}
 \caption{The architecture of the RRCN model.}
 \label{fig:RRCN}
\end{figure}

\textbf{\textit{RRCN}}: \textcolor{blue}{Li et al. 2019} presented a very deep non-simultaneous fully recurrent convolutional network for VSR. Specifically, they applied very deep fully recurrent convolutional layers and late fusion on motion compensated video frames to make full use of the temporal information in their non-simultaneous recurrent convolutional network architecture. A CLG-TV optical flow method (\textcolor{blue}{Drulea and Nedevschi 2011}) is utilized for motion estimation, and the centering frame is chosen as the reference frame to compensate the adjacent frames, then both the centering frame and the motion compensated frames are fed into their network for VSR. Remarkably, the very deep recurrent convolutional network has powerful representation ability. Besides, it is good at modeling the spatial and temporal non-linear mappings. The architecture of the RRCN model is depicted in Figure~\ref{fig:RRCN}.

\begin{figure}[!h]
 \centering
 \includegraphics[width=0.75\linewidth]{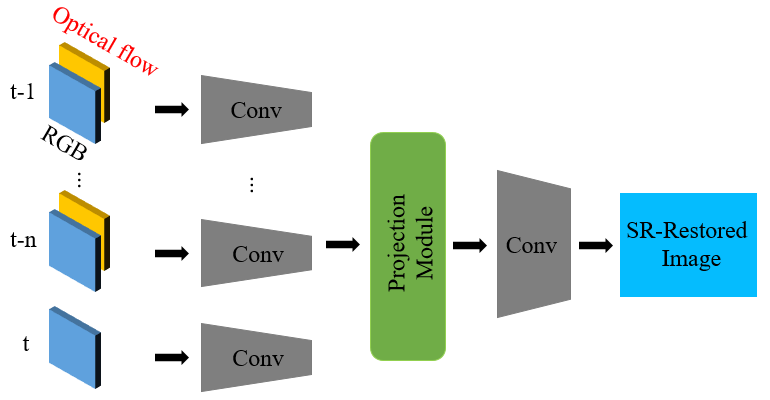}
 \caption{The architecture of the RBPN model.}
 \label{fig:RBPN}
\end{figure}

\textbf{\textit{RBPN}}: \textcolor{blue}{Haris et al. 2019} proposed a Recurrent Back-Projection Network (RBPN), which consists of three main components: a feature extraction module, a projection module, and a reconstruction module. RBPN collects the spatial and temporal information from video frames surrounding the target one. Back-projection in the recurrent process is used to organize the temporal information, where high-resolution features can be gradually refined and applied to reconstruct the high resolution target frame. Optical flow is used for initial feature extraction. Specifically, before entering projection modules, they contact the pre-computed optical flow motion maps with the target frame $I_t$ in the corresponding neighborhood $[I_{t-N}, …, I_{t-k}, …, I_t]$, \textit{N} is the temporal neighborhood frame numbers. The optical flow motion map can encourage the projection module to capture missing details between $I_t$ and its neighbors $I_{t-k}$. For the reconstruction part, DBPN (\textcolor{blue}{Haris et al. 2018}) is employed as the single image super-resolution network, and ResNet (\textcolor{blue}{He et al. 2016}) with deconvolution is utilized as the multi-image super-resolution network. The architecture of the RBPN model is depicted in Figure~\ref{fig:RBPN}.

\textbf{\textit{Limitation}}. RNN based VSR methods generally suffer from the following limitations:
\begin{itemize}
\item It is good at modelling global slow-varying motions but not those short-term fast-varying ones. This is because the recurrent connections operate on hidden states, while significant fine-grained details for depicting fast-varying motions mostly exist in input video frames other than the hidden states (Baker and Kanade 1999; Huang et al. 2018).

\item Without explicit temporal alignment, the RNN based VSR methods have limited ability to deal with complex and large motions.

\item The structure information, which is important for super-resolving LR video frames, is lost. This is caused by the dimensionality reduction when transforming the input 2D LR video frames to the 1D vectors of RNN hidden states.
\end{itemize}

\begin{landscape}
\begin{table}[!htp]
\caption{Comparison of CNN (+ Optical Flow) based VSR Methods. MEC denotes Motion estimation \& compensation.}
\label{tab:FlowModelComp}
\begin{tabular}{llll}
\hline
\textbf{Method} & \textbf{Year}  & \textbf{MEC} &  \textbf{Advantage} \\
\toprule[1pt]   
\textbf{Deep-DE} &  2015  &  Explicit MEC, Optical flow & \makecell[l]{1) SR draft ensemble generation: Renovate traditional feed-forward reconstruction pipeline; \\Enhance ability to compute different super-resolution results; \\Consider large motion variation and various latent artifacts.
\\ 2) SR draft based optimal reconstruction: \\Using CNN to integrate the reconstruction and deconvolution steps; \\Avoiding parameter tuning in the test phase.} \\
\specialrule{0em}{3pt}{3pt}
\textbf{VSR-Net} & 2016  &  Explicit MEC, Optical flow & \makecell[l]{1) Proposed a CNN trained on both the spatial and temporal spaces to enhance the spatial resolution; \\2) Integrating motion compensated frames, filter symmetry enforcement, and pre-training strategy \\to improve both the accuracy and efficiency; \\3) Exploited an adaptive motion compensation method to handle motion blur and fast moving objects.} \\
\specialrule{0em}{3pt}{3pt}
\textbf{VESPCN} & 2017  &  Explicit MEC, Optical flow & \makecell[l]{1) Studying different fusion approaches to discover spatio-temporal correlations; \\2) Building a motion compensation scheme based on a fast multi-resolution spatial transformer module; \\3) Exploiting a spatio-temporal sub-pixel convolution network to improve accuracy and temporal \\consistency of VSR in real-time.} \\
\specialrule{0em}{3pt}{3pt}
\textbf{STTN} & 2018  &  Explicit MEC, Optical flow & \makecell[l]{1) Exploring a spatio-temporal flow estimation network to selectively capture long-range temporal dependency \\efficiently and handle occlusion effectively;
\\2) Presenting a spatio-temporal sampler to enable spatio-temporal manipulation;
\\3) Integrating on the top of conventional networks easily.} \\
\specialrule{0em}{3pt}{3pt}
\textbf{TOFlow} & 2019  &  Explicit MEC, Optical flow & \makecell[l]{1) Proposed a TOFlow network to learn flow representation in a self-supervised, task-specific way; \\2) Jointly learn the task-oriented optical flow and perform VSR; \\3) The network is end-to-end trainable.}\\
\specialrule{0em}{3pt}{3pt}
\textbf{MMCNN} & 2019  &  Explicit MEC, Optical flow & \makecell[l]{1) Designing a MMCNN for accurate and fast VSR by cascading an optical flow module and an \\image-reconstruction module; \\2) Replacing sub-pixel motion compensation with motion transformer operator to faster optical flow estimation; \\3) Embedding LSTM into the residual block to form a multi-memory residual block to progressively learn \\and retain temporal dependency between adjacent LR frames; \\4) Using a series of residual blocks engaged in terms of intra-frame spatial correlations for feature extraction \\and reconstruction.} \\
\hline
\end{tabular}
\end{table}
\end{landscape}

\begin{landscape}
\renewcommand\thetable{1}
\begin{table}[!htp]
\caption{(Continued).}
\label{tab:FlowModelComp}
\begin{tabular}{llll}
\hline
\textbf{Method} & \textbf{Year}  & \textbf{MEC} &  \textbf{Advantage} \\
\toprule[1pt]   
\textbf{SoSR-ToSR} & 2019  &  Explicit MEC, Optical flow & \makecell[l]{1) Designing a two-stream network for VSR; \\2) Proposing SoSR with optical flow guided weighted MSE loss to pay more attention to moving objects; \\3) Exploiting ToSR with a siamese network to emphasize temporal consistency.}  \\
\specialrule{0em}{2.5pt}{2.5pt}
\textbf{MEMC-Net} & 2019 &  Explicit MEC, Optical flow & \makecell[l]{1) Proposed a motion estimation and compensation driven neural network for robust and high-quality \\video frame interpolation; \\2) Exploited an adaptive warping layer to integrate both optical flow and interpolation kernels to \\synthesize target frame pixels; \\3) This adaptive warping layer is fully differentiable to enable to jointly optimize the flow and kernel \\estimation networks.} \\
\specialrule{0em}{2.5pt}{2.5pt}
\textbf{MultiBoot-VSR} & 2019 &  Explicit MEC, Optical flow & \makecell[l]{1) Proposed a scene and class agnostic, fully convolutional neural network; \\2) The model consists of a motion compensation based input subnetwork, a blending backbone, and \\a spatial upsampling subnetwork; \\3) Reusing reconstructed high-resolution frames as additional reference frames after reshuffling them \\into multiple low-resolution images to bootstrap and enhance image quality progressively.} \\
\specialrule{0em}{2.5pt}{2.5pt}
\textbf{SOF-VSR} & 2020 &  Explicit MEC, Optical flow & \makecell[l]{1) Designed a unified SOF-VSR to jointly super-resolve optical flow and images; \\2) Exploited an OFRnet to infer HR optical flows from LR frames in a coarse-to-fine manner to \\recover accurate temporal details for SR; \\3) Motion compensation is conducted via HR flows to encode temporal dependency.} \\
\specialrule{0em}{2.5pt}{2.5pt}
DRVSR & 2017 &  Explicit MEC, Optical flow & \makecell[l]{1) Can take arbitrary-size input images; \\2) Exploited a SPMC to better handle inter-frame motion; \\3) SPMC layer can be used for arbitrary scaling factors during testing as it without trainable parameters; \\4) A DF network is applied to effectively fuse image details from multiple images after SPMC alignment.} \\
\specialrule{0em}{2.5pt}{2.5pt}
TANN & 2018 &  Explicit MEC, Optical flow & \makecell[l]{1) Proposed a DDAN for VSR which includes a MCNet and a ReconNet to fully exploit the spatio-temporal \\informative features; \\2) MCNet progressively extracts flow representations to synthesize the motion information across adjacent \\frames in a pyramid fashion; \\3) Capturing detail components of original LR adjcent frames as complementary cues for accurate feature \\extraction to reduce mis-registration errors of motion estimation; \\4) Conducting dual attention on a residual unit and form a residual attention unit to emphasize meaningful \\features for high-frequency details recovery.} \\
\hline
\end{tabular}
\end{table}
\end{landscape}

\begin{landscape}
\renewcommand\thetable{1}
\begin{table}[!htp]
\caption{(Continued).}
\label{tab:FlowModelComp}
\begin{tabular}{llll}
\hline
\textbf{Method} & \textbf{Year}  & \textbf{MEC} &  \textbf{Advantage} \\
\toprule[1pt]   
\textbf{STR-ResNet} & 2018 &  Implicit MEC, RNN & \makecell[l]{1) Explored a STR-ResNet VSR method by jointly modeling intra-frame redundancy and inter-frame \\motion context in a unified deep neural network; \\2) The STR-ResNet is the first study attempts to incorporate the bypass connection in a deep network \\to embed the joint spatial-temporal residue prediction and model temporal correlations in video frames; \\3) The network is able to implicitly model the motion context among multiple video frames for VSR; \\4) It among the first to investigate and integrate the spatial convolutional, temporal recurrent and \\residual architectures into a single deep network to address VSR.} \\
\specialrule{0em}{3.5pt}{3.5pt}

\textbf{FRVSR} & 2018 &  Implicit MEC, RNN & \makecell[l]{1) Exploited an end-to-end trainable frame-recurrent VSR method by using the previously inferred HR \\estimate to super-resolve the subsequent video frame; \\2) Ensuring temporally consistent and reducing the computational cost by warping only one image in \\each step; \\3) Enabling to assimilate a large number of prior frames without increased computational requirement; \\4) Without needing any pre-training stages.} \\
\specialrule{0em}{3.5pt}{3.5pt}

\textbf{RBPN} & 2019 &  Implicit MEC, RNN & \makecell[l]{1) A RBPN is proposed to treat each frame as a separate source, the sources are integrated in an \\iterative refinement framework via back-projection modules; \\2) Integrating SISR and MISR in a unified VSR framework; \\3) Explicitly representing estimated inter-frame motion with respect to the target instead of explicitly \\aligning frames.} \\
\specialrule{0em}{3.5pt}{3.5pt}

\textbf{RRCN} & 2019 &  \makecell[l] {Explicit MEC, Optical flow, \\ RNN (Model temporal dependency)} & \makecell[l]{1) Proposed the first very deep non-simultaneous fully RNN for VSR; \\2) A model ensemble method is exploited to combine multi-frame SR model with single-image SR \\model; \\3) Strong representation ability; \\4) Good at modeling the spatial and temporal non-linear mappings.} \\
\hline
\end{tabular}
\end{table}
\end{landscape}

\section{Challenge and Future Direction}
\label{sec:12}
Although great progress has been made for VSR in the past decades, there remain some open research questions. In this section, we will discuss these challenges explicitly and outline some promising directions for future study. The challenges and trends are investigated in two aspects.

\subsection{Challenges for optical flow-based temporal alignment}
\label{sec:13}
Since accurate optical flow estimation remains challenging for real videos, the optical flow-based temporal alignment is still a key problem in VSR, where the main challenges including:
\begin{itemize} 
\item Inaccurate flow will result in distortion and errors, deteriorating the final VSR performance.
\item Image-level warping strategy introduces artifacts into the aligned frames.
\item Fast moving and large displacement are difficult to handle, which significantly affect the performance of both optical flow estimation and flow warping.
\item Per-pixel motion estimation suffers a heavy computational cost.
\end{itemize}

\subsection{Future direction for video super-resolution}
\label{sec:14}
\subsubsection{Lightweight VSR Model}
Currently, most VSR methods emphasize on pursuing high performance, leading to large models (contain a huge number of parameters) that take a long time for training while require high computing and storage resources. These characteristics prevent their use on mobile devices in practical applications, where small models and fast inference speed are preferred. Therefore, how to design lightweight VSR models while still maintaining high performance is a promising research topic.

\subsubsection{Applicable to Real-world Scenarios}
\label{sec:15}
VSR methods face difficulties in real-world scenarios as they often suffer from issues like unknown degradation, scene change, occlusion, etc. Boosting the ability of VSR methods to handle such real-world scenarios is urgent.
\begin{itemize} 
\item \textbf{Handling Degradation}. Existing VSR methods normally generate LR video frames according to the manners of downsampling directly with interpolation (e.g. bicubic interpolation) or downsampling after Gaussian blurring, manually. However, it is well-known that the real-world degradation process is very complicated which includes many uncertainties. As a result, VSR models, which are trained on artificially produced degradation by blurring and interpolation, cannot well conformed to actual LR video frames in practice. Consequently, better ways of modeling and handling degradation are needed.  
\item \textbf{Handling complex scene changes}. In reality, a video often contains some different scenes. However, the current VSR methods cannot deal with such changes well. A typical approach is split the videos into multiple segments, each without scene changes, and then process them individually. This kind of strategy increases the computational cost, therefore new ways to handle videos with scene changes are necessary for realistic VSR applications.
\end{itemize}

\subsubsection{Effectively Exploring Spatio-temporal Information}
Except the appearance RGB information, video includes the temporal information. How to effectively explore temporal information across video frames will directly affect the performance of VSR. Current methods, like 3D convolution and non-local modules are computationally inefficient, and the quality of optical flow cannot be guaranteed. Consequently, proposing new methods to effectively make use of spatio-temporal information in video is worth further study.

\subsubsection{More Reasonable Evaluation Metrics}
Evaluation metric is the most fundamental component in each computer vision task, which greatly influences the task’s progress. Exploring more reasonable evaluation metrics is of equal importance to exploit more advanced algorithms.

Similar to image SR, nowadays, the performance of video SR is also evaluated mainly by the FR-IQA metrics like peak signal-noise ratio (PSNR) and structural similarity index (SSIM) (\textcolor{blue}{Wang et al. 2004; Yang et al. 2019}). These FR-IQA measures face some fatal challenges: (1) They are unable to reflect the video quality for human perception quite well. Because these methods depend on the pixel-level error measures, like L1 and L2 distances or their combination (\textcolor{blue}{Timofte et al. 2018}), causing them concentrate on local pixel-level information, thus they cannot measure perceptual quality accurately (\textcolor{blue}{Ledig et al. 2017b; Ma et al. 2017}). \textcolor{blue}{(Blau et al. 2018)} has demonstrated that images with high PSNR and SSIM produce overly smooth images with low perceptual quality; (2) They are designed in a limited and refined condition, which require manual intervention, thus they lack of feasibility for modeling unknown distortions \textcolor{blue}{(Anwar et al. 2020)}. (3) They require non-distorted ground-truth images for comparison, which are almost unavailable in practice, because it is difficult or impossible to acquire ideal reference images in most conditions, especially for the real video data.

To address this issue, some new perceptual-based NR-IQA metrics have been proposed (\textcolor{blue}{Kim and Lee 2017; Talebi and Milanfar 2018; Zhang et al. 2018b; Prashnani et al. 2018; Tang et al. 2019; Zhu et al. 2020}). However, there are still no universally accepted evaluation criteria that can work in various situations and perfectly assess SR quality. Even worse, we do not make clear what kind of perceptual quality is real important and useful for assessing SR currently. Nevertheless, proposing new effective metrics which can be broadly used is remaining an open research problem.

\subsubsection{Unsupervised VSR}
The state-of-the-art VSR approaches are deep neural network based and trained in the supervised manner. However, deep learning networks require a large number LR-HR video frame pairs for training. On the one hand, the paired datasets are rare or costly to acquire in practice. On the other hand, when the input video frames are poor in resolution, the super-solution cannot work well. Furthermore, the current VSR models trained on these artificial labeled datasets can only learn the inverse process of the predefined degradation, which is too simple to characterize the real-world situation. One promising direction is to exploit unsupervised VSR methods which can be well performed on unpaired LR-HR video sets.

\section{Conclusion}
\label{sec:Conc}
Video super-resolution (VSR), which aims to improve the clarity and visual appearance of video frames, is a crucial task in computer vision and has been deeply investigated. One core problem for VSR is how to capture temporal dependency to achieve efficient and accurate temporal alignment. Noticeably, the most popular way is to use optical flow to acquire motion information for explicit or implicit temporal alignment. In this work, we provide a comprehensive review of optical flow based VSR methods by introducing previous work and analyzing current advances, and discussing their limitations incidentally. In particular, we firstly explain what is video super-resolution. Secondly, we give a detailed explanation about what is optical flow based VSR. Thirdly, both the representative traditional (i.e. reconstruction based VSR method and learning based VSR method) and the current deep learning based VSR algorithms that make use of optical flow are compared and explored. Remarkably, we deeply investigate the deep neural network related VSR methods. Fourthly, we find that although deep learning based VSR algorithms have achieved great progress, there are still some practical issues and unsolved problems. Accordingly, we discuss the challenges and point out the promising future research trend for VSR. To the best of our knowledge, this is the first systematical work on surveying the effect of optical flow in VSR. We hope this survey would not only provide a better and deeper understanding of optical flow based VSR, but also serve as a catalyst to spur future research activities in this domain.

To promote the future research activity, we hope to play as a forerunner and start the discussion from the following aspects:

1) \textbf{\textit{Lightening VSR Model with Knowledge Distillation}}. There is a considerable performance gap between the lightweight VSR model and the normally used complex VSR model, while the latter one requires a much larger amount of resources (\textcolor{blue}{Xiao et al. 2021}). This problem is particularly acute on resource-limited devices, e.g., smartphones and wearable devices. Where a compact VSR model can be easily used on these devices, but due to its limited capacity to model spatial-temporal correlations, the VSR performance is unsatisfactory. Knowledge distillation, which is able to transfer knowledge from a complicated model (teacher network) to a simplified one (student network), and without altering the original architecture of the teacher network, supplies a possible way to handle this problem. Designing a spatial-temporal distillation (STD) scheme, which is suitable for VSR, is a promising research topic.

\textbf{\textit{2) Promoting VSR for Real-world Scenarios}}. Although VSR methods have achieved remarkable progress recently, they are still unapplicable in reality. One reason is that most of the existed VSR models are trained and assessed on synthetic datasets, where the videos are generated by simple synthetic degradation methods. Badly, these degradation methods unable to well simulate the complicated degradation processes in realistic videos, and accordingly leading to the trained VSR models noneffective for real-world applications (\textcolor{blue}{Yang et al. 2021}). Research Topic 1: building a real-world video super-resolution dataset, which can bridge the synthetic-to-real gap in VSR and supply a valid benchmark for training and evaluating the real-world VSR models generally. The second reason is that the prior degradation models take some factors into account, e.g. blur, downsampling, noise, but they are still cannot cover the diverse degradations of real video data (\textcolor{blue}{Pan et al. 2021}). Research Topic 2: exploiting a practical degradation model that consists of random degradations. For example, designing a VSR which can simultaneously estimate unknown blur kernels, motion fields, and latent HR videos effectively is prospective. Lots of VSR models have been proposed, but there is not a unified framework being dominant for VSR in practice yet (\textcolor{blue}{Chan et al. 2021; Yi et al. 2021}). Research Topic 3: Exploring a generic, efficient, and easy-to-implement baseline framework for VSR, which can serve as a standard for various comparison and evaluation.

3) \textbf{\textit{Exploiting More Useful Spatio-temporal Information}}. Capturing spatio-temporal information accurately and efficiently is critical important for VSR. Recently, using the deformable convolution backbone to conduct spatio-temporal VSR on the feature space directly is popular as this strategy is fast (\textcolor{blue}{Xiang et al. 2020}). However, these VSR models would only produce pre-defined intermediate frames, causing them constrained to highly-controlled scenarios with fixed frame-rate videos. Consequently, exploiting controllable spatio-temporal VSR approaches, which with the deformable convolution network, for smooth motion synthesizing it is necessary (\textcolor{blue}{Xu et al. 2021}).

In addition, current VSR models underrated the short-term motion cues between successive video frames, therefore how to exploit both the short-term and long-term motion cues in videos is desirable.

4) \textbf{\textit{Designing Natural Evaluation Metrics}}. Since the ground truth reference image is almost absent for real data, it means that the widely used FR-IQA metrics are unreasonable in fact, human evaluation is the only rational way to evaluate the performance of VSR models. Mean-Opinion-Score (MOS), i.e. the average rating that human raters assigned to super-resolved images via a certain SR model, is the typical human evaluation measure. However, the MOS values of different models are not directly comparable due to the changing of rater numbers and rater’s subjectivity, etc (\textcolor{blue}{Khrulkov and Babenko 2021}). Consequently, it is urgent to propose new NR-IQA metrics, which enable to break through the current evaluation predicament in some aspects: (1) comparing with various VSR models automatically, (2) approximating human preferences naturally; (3) tuning hyperparameters without artificial assistance effectively, etc.

\textbf{\textit{5) Improving Unsupervised VSR}}. For unsupervised VSR, there are main two ways. First, super-resolving video images without introducing predefined degradation by using unpaired LR-HR datasets. Despite great progress, it is hard to synthesize good ``real" LR images for super-resolution yet (\textcolor{blue}{Wang et al. 2021b}). Formulating unpaired SR training as a domain adaptation issue, and enabling the VSR network to match LR images from different domains into a shared degradation-imperceptible feature space deserved to be discussed. Second, learning model with unsupervised deep networks, where Generative Adversarial Network (GAN) is the currently primary method. However, GAN usually brings noise and causes some details of dislocation. Exploring more advanced unsupervised deep networks is another research topic.

\begin{acknowledgements}
This work was supported by the National Natural Science Foundation of China under Grant 62106177. It was also supported by the Central University Basic Research Fund of China (No.2042020KF0016, CCNU20TS028), the Teaching research project of CCNU (202013), and the Wuhan University-Infinova project No.2019010019. The numerical calculation was supported by the supercomputing system in the Super-computing Center of Wuhan University.
\end{acknowledgements}

\section*{Conflict of interest}
The authors declare that they have no conflict of interest.


\end{document}